\documentclass{article}
\PassOptionsToPackage{numbers, compress}{natbib}

\usepackage[preprint]{neurips_2020}
\usepackage{graphicx}

\usepackage[utf8]{inputenc} 
\usepackage[T1]{fontenc}    
\usepackage{xr-hyper} 
\usepackage{hyperref}       
\usepackage{url}            
\usepackage{booktabs}       
\usepackage{amsfonts}       
\usepackage{nicefrac}       
\usepackage{microtype}      

\usepackage{amsmath}

\title{Efficient Representation of Natural Image Patches}

\author{%
  Cheng Guo \\
  \texttt{cheng.guo.work@gmail.com} \\
}

\begin{document}

\maketitle

\begin{abstract}
    Utilizing an abstract information processing model based on minimal yet realistic assumptions inspired by biological systems, 
    we study how to achieve the early visual system's two ultimate objectives: efficient information transmission and accurate sensor probability distribution modeling. 
    We prove that optimizing for information transmission does not guarantee optimal probability distribution modeling in general. 
    We illustrate, using a two-pixel (2D) system and image patches, 
    that an efficient representation can be realized through a nonlinear population code 
    driven by two types of biologically plausible loss functions that depend solely on output. 
    After unsupervised learning, our abstract information processing model bears remarkable resemblances to biological systems, 
    despite not mimicking many features of real neurons, such as spiking activity. 
    A preliminary comparison with a contemporary deep learning model suggests that our model offers a significant efficiency advantage. 
    Our model provides novel insights into the computational theory of early visual systems 
    as well as a potential new approach to enhance the efficiency of deep learning models.
\end{abstract}

\section{Introduction}
Biological neural systems are intricate masterpieces of nature's information processing mechanisms \citep{Sanes2010, Masland2012}. 
Just as airplanes achieve flight without mimicking the exact structure of feathers, 
some intricacies of the biological neural system might arise from biological constraints than from fundamental information processing requirements \citep{barlow1961possible, Marr}.
Despite extensive research and significant advancements, 
discerning the essentials from the ancillary when constructing a computational theory for neural systems remains a challenge \citep{Carandini2005}. 
Theoretical models aiming to capture detailed aspects of neural systems often rely on many assumptions and approximations 
\citep{dayan2005theoretical, izhikevich2004model, marder2011multiple}. 
This may limit their applicability or neglect other important aspects.

Artificial neural networks and contemporary deep learning methodologies 
have been profoundly influenced by their biological counterparts \citep{Hassabis2017, lecun2015deep}. 
Yet, their primary objective is to solve specific tasks, with their merit being assessed predominantly based on their effectiveness.
As a result, the question of whether these methods are principled or reflect crucial features of biological systems is often sidelined or deemed irrelevant.
A notable illustration of this is the concept of efficient coding \citep{barlow1961possible, attneave1954some, SIMONCELLI2003144}.
Despite being a cornerstone in computational neuroscience grounded in information theory and having ample experimental support, 
it hasn't been a central theme in the design principles of artificial neural networks.

In this paper, we address the aforementioned problems using an ab initio approach.
First, we identify the fundamental assumptions of an abstract discrete feedforward information processing unit model amidst the myriad features of biological systems.
Next, we examine this abstract model, starting from a single pixel and progressively extending to image patches, 
and then compare the results with those of both biological systems and deep learning models. 
Through this approach, we aim to offer novel insights into the computational theory of early visual systems 
and aspire to enhance the efficiency of deep learning models.

\section{Assumptions}

We define the objectives, the means, and the characteristics of a discrete feedforward information processing unit (DFIPU or IPU for short) with the following assumptions:

\paragraph{Assumption A:} \textit{An IPU strives to accomplish two ultimate objectives: information transmission and input probability distribution modeling.}

To optimize survival, organisms need to accurately and efficiently relay new information throughout their systems for processing and responses. 
Furthermore, they benefit from predicting environmental occurrences, or in mathematical terms, understanding the probability distribution of their environment, 
based on both personal experiences and inherited evolutionary memory.

The first goal aligns with the efficient coding hypothesis and has been extensively studied by the computational neuroscience community 
\citep{Linsker1988, van1992theory, Atick1992, simoncelli2001natural, karklin2011efficient}.

The second goal, modeling the input probability distribution, has been explored in various contexts using different methods
such as principal component analysis (PCA) \citep{hancock1992principal, tipping1999probabilistic},  
Bayesian networks \citep{jensen2007bayesian},
Markov random fields (MRF) \citep{geman1984stochastic, Osindero2008}, 
independent component analysis (ICA) \citep{bell1997independent, van1998independent}, sparse coding \citep{olshausen1997sparse},
distribution coding model for complex cells \citep{Karklin2009},
restricted Boltzmann machines (RBM) \citep{Hinton2006}, products of experts (PoE) \citep{hinton2002training}, 
field of experts (FoE) \citep{roth2005fields}, Gaussian mixture model (GMM) \citep{zoran2011learning}, 
variational autoencoders (VAE) \citep{kingma2013auto} and generative adversarial networks (GAN) \citep{goodfellow2014generative} etc.

There are also a few works investigating the relationship between the two goals \citep{Cardoso1997, olshausen1997sparse, ganguli2014efficient}. 
Efficient coding leads to solutions that allocate more resources to represent input regions with high probability density. 
This might give the impression that the two goals are equivalent.
What adds to the confusion is the fact that for the widely used ICA, the two objectives have indeed been proven to coincide \citep{Cardoso1997}.
Additionally, the seminal sparse coding loss function can be derived, given certain reasonable assumptions and approximations, 
either from the efficient coding perspective \citep{Olshausen1996} or the input probability modeling perspective \citep{olshausen1997sparse}.

With our IPU model, we prove the two goals are distinct in general.
We will also compare different pragmatic strategies to achieve these two goals.

\paragraph{Assumption B:} \textit{Both the input and output of the IPU are discrete, 
with the number of output states N being significantly fewer than the number of input states M ($M \gg N$).}

On the output end, neurons produce either spikes or graded outputs. Spikes are discrete events. 
Rate coding, therefore, offers limited resolution within a predetermined time frame. 
In temporal coding, the timing of spikes also has finite resolution \citep{bair1996temporal, Butts2007}. 
For graded outputs, biological constraints limit the resolution. 
For instance, luminance resolution levels at synapse terminals in a zebrafish's retina are just around 10 \citep{Odermatt2012}.

On the input end, those neurons receiving inputs from other neurons, consequently possess finite input states.
For sensory neurons, external signals undergo conversion into biochemical signals, which due to molecular constraints, have finite resolution.
Moreover for visual signals, the quantization of light into photons by quantum mechanics also establishes an ultimate constraint on resolution.

When neurons receive inputs from others, it is evident that $M \gg N$ given that $M \sim N^L$, where L is the number of inputs.
For sensory neurons, as illustrated in the zebrafish example, the relationship $M \gg N$ holds true.
Even if, for some sensory neurons in certain species, the relationship $M \gg N$ is not evident,
we can still group both the sensory neurons and the neurons receiving their outputs into a single IPU model, ensuring $M \gg N$ holds.
With $M \gg N$, information passing through each IPU undergoes significant compression through clustering and categorization, 
setting the stage for subsequent processing tasks.

\paragraph{Assumption C:} \textit{In the limit as $M \to \infty$, the input becomes continuous; 
in the limit as $N \to \infty$, the IPU transformation becomes a continuous function, resulting in continuous output.}

Previous theoretical studies utilizing tools such as calculus naturally assume that inputs are continuous and that transformations are continuous functions \citep{van1992theory, Atick1992, Bell1995, ganguli2014efficient}. 
This assumption stems from the fundamental continuity of an organism's natural environment. 
Moreover, without continuity, models would be unable to make inferences based on the immediate vicinity of data points or to learn effectively.

Although this study assumes discrete inputs, we want to retain certain degree of "continuity". 
We conceptualize the discrete input values as the results of a quantization process applied to an underlying continuous quantity, 
with the resolution of this process defined by $M$. Hence, as $M \to \infty$, the discrete input converges towards the underlying continuous quantity. 
Similarly, we envisage the transformation as a continuous function, and the output as continuous, when the output resolution $N \to \infty$. 
This assumption also enables the stacking of IPU models for multi-stage processing.

\paragraph{Assumption D:} \textit{The IPU transformation is deterministic.}

For simplicity in this study, we examine only the noiseless case where the IPU transformation is deterministic. 
We argue that this is a good approximation because responses of neurons in the early visual system 
to repeated stimuli have been found to be highly reproducible  \citep{mainen1995reliability, Berry1997}.


\section{One Pixel}
We begin with the simplest case, where the input to the model consists of a single pixel with one color channel. 
Although this model might seem trivial, it can represent various biological entities. 
For instance, it could model the eyespot of single-celled organisms such as Euglena, 
the large monopolar cells found in an insect’s compound eye, or the bipolar cells in the retina. 
The single pixel case has been studied extensively \citep{Laughlin1981,Atick1992, Bell1995, ganguli2014efficient}. 
We aim to revisit it to introduce the concepts and notation.

Let us denote the light intensity of the pixel as $x \in X$, 
where $X = \{x_1, x_2, \ldots, x_M\}$ represents a set of discrete input states. 
Let $p(x)$ represent input probability distribution, satisfying $\sum_x p(x) = 1$. 
The information of $x$ is quantified by Shannon's entropy:
\begin{align}
H_p = -\sum_{i=1}^M p(x_i) \log{p(x_i)}.
\end{align}

\begin{figure}
    \centering
    \begin{tabular}{ccc}
      \includegraphics[width=0.3\textwidth]{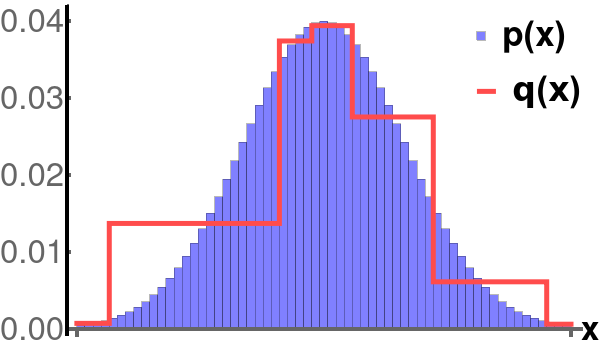} & \includegraphics[width=0.3\textwidth]{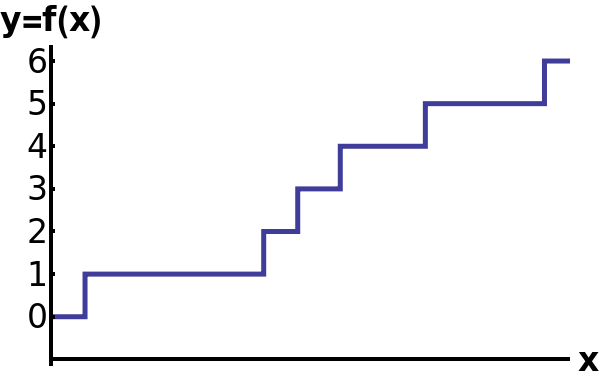} & \includegraphics[width=0.3\textwidth]{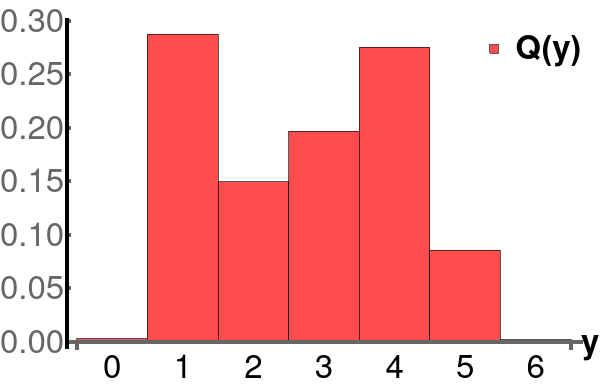} \\
      (a) & (b) & (c) \\ [6pt]
    \end{tabular}
    \caption{Example illustrating how an IPU models input probability $p(x)$:
    (a) This panel shows the example input probability distribution $p(x)$ alongside the approximation learned by the IPU model, $q(x)$, calculated using Eq.~(\ref{eq:qx2}). 
        Here, $x$ is discrete and has $M$ different states.
    (b) The many-to-one function $y = f(x)$ categorizes the input states into $N=7$ distinct groups within the output space.
    (c) With both $p(x)$ and $f(x)$ given, one can calculate the output distribution $Q(y)$ 
        over the $N=7$ output states using Eq.~(\ref{eq:Qy}).}
\label{fig:px_qx}
\end{figure}

Under Assumptions B and D, the IPU deterministically maps the input $x$ to the output $y$ through a many-to-one transformation function, defined as
\begin{align}
f: X \rightarrow Y, \quad &X = \{x_1, x_2, \ldots, x_M\}, \quad Y = \{y_1, y_2, \ldots, y_N\}, \quad \text{where } M \gg N.
\label{eq:y}
\end{align}
This mapping function behaves analogously to a step function by mapping multiple distinct inputs from $X$ to a single output in $Y$ 
across defined groups within $X$ (see Fig.~\ref{fig:px_qx}(b) for an example). 
Each group $j$, denoted as $G_j$, contains $x$ values with a group size of $n_j$. 
It holds that $\sum_j^N n_j = M$. 

The entropy of the output $y$ is given by
\begin{align}
    H_Q = - \sum_{j=1}^N Q(y_j) \log Q(y_j),
\label{eq:HQ}    
\end{align}
where $Q(y_j)$ represents the probability distribution of the output states. By definition
\begin{align}
    Q(y_j) = \sum_{x \in G_j}p(x).
\label{eq:Qy}
\end{align}
In this paper, we use a uppercase letter to denote the distribution over the output states, and a lowercase letter for the distribution over the input states.

To optimize the goal of information transmission, we need to maximize the rate of transmission, or the mutual information between input and output \citep{barlow1961possible, Linsker1988}.
Many previous studies model a neuron's spike activity as a probabilistic event, which renders the mapping from stimuli to responses probabilistic. 
In these cases, mutual information is difficult to compute, and researchers often have to resort to approximations \citep{karklin2011efficient, ganguli2014efficient, wei2016mutual}.
However, our abstract IPU model is deterministic (Assumption D), and the mutual information simply equals $H_Q$ (see proof in Appendix \ref{appendix:rot}).
Maximizing $H_Q$ instead of mutual information is easier to deal with both analytically and numerically.

Simultaneously, an IPU should also strive to fulfill the second objective: modeling $p(x)$.
Given $p(x)$ and $y=f(x)$ one can determine $Q(y)$ using Eq.~(\ref{eq:Qy}).
Conversely, given $Q(y)$ and $y=f(x)$ it is not possible to perfectly recover the original $p(x)$;
instead, one can only approximate $p(x)$ with a probability $q(x)$ due to the many-to-one mapping from $x$ to $y$ (see Fig.~\ref{fig:px_qx}).
The approximation $q(x)$ satisfies the following relation for every $G_j$
\begin{align}
    \sum_{x \in G_j}q(x) = \sum_{x \in G_j}p(x).
\label{eq:px_qx}
\end{align}
We can calculate the approximation $q(x)$ from $Q(y)$ with $q(x) = \sum_j p(x|y_j) Q(y_j)$.
For the deterministic step function $y=f(x)$, the conditional probability $p(x|y_j)$ is given by
\begin{align}
    p(x|y_j)= 
\begin{cases}
    1/n_j, & \text{if } x \in G_j\\
    0,              & \text{otherwise}.
\end{cases}
\label{eq:p_x_y}
\end{align}
This leads to
\begin{align}
    q(x) = 
    q_j = \frac{Q(y_j)}{n_j}, \quad  \text{for } x \in G_j.
\label{eq:qx2}
\end{align}

The IPU model does not output $q(x)$ explicitly but rather encodes $q(x)$ in the transformation $y=f(x)$.
This aligns with the findings of Ganguli and Simoncelli \citep{ganguli2014efficient},
who also demonstrated how such an implicitly encoded $q(x)$ can be extracted and utilized for inference.
In this paper, we concentrate on representation learning, leaving the topic of inference for future studies.

Mathematically, achieving the second goal entails minimizing the Kullback–Leibler divergence (KL-divergence) between $p(x)$ and $q(x)$.
Next, we investigate the relationship between the two goals using our IPU model and determine whether an optimal solution for one leads to the optimal solution for the other.

\section{Relation Between the Two Goals}
\label{section:two_goals}


We can calculate KL-divergence between $p(x)$ and $q(x)$ using
\begin{align}
    D_{KL}(p||q) = H_{pq} - H_p,
\end{align}
where $H_{pq}$ is the cross entropy. It can be proved that the cross entropy $H_{pq}$ is equal to the entropy of $q(x)$
defined as (see proof in Appendix~\ref{appendix:cross}):
\begin{equation}
    H_q = -\sum_{i=1}^M q(x_i) \log q(x_i).
    \label{eq:Hq}
\end{equation}
Therefore we have
\begin{equation}
    D_{KL}(p||q) = H_q - H_p.
\label{eq:KL-divergence}
\end{equation}
Since $H_p$ is fixed, minimizing the KL-divergence requires minimizing $H_q$. 

To understand the relation between the two goals is to understand the relation between 
maximizing the output entropy $H_Q$ and minimizing the entropy of the modeled distribution $H_q$.
Let us find out how $H_q$ would change while we maximize $H_Q$.

Consider two adjacent zones in the transformed space, wherein the corresponding probabilities $Q(y_1)$ and $Q(y_2)$ differ,
let us assume $Q(y_1) > Q(y_2)$. 
To maximize $H_Q$, one can reduce the inequality by shifting the boundary between these two zones and 
moving one $x$ value from $G_1$ to $G_2$. 
Under Assumptions B and C, this shift equates to a minor alteration in the probabilities for both zones, denoted by $\delta$,
resulting in $Q(y_1)$ adjusting to $Q(y_1) - \delta $ and $Q(y_2)$ to $Q(y_2) + \delta$.

In the meantime, to calculated the change of $H_q$ we first combine Eq.~(\ref{eq:qx2}) and Eq.~(\ref{eq:Hq}) to get 
\begin{align}
    H_q 
    &=  -\sum_j \sum_{x \in G_j} q(x) \log q(x) \\
    &= -\sum_j Q(y_j) \log \frac{Q(y_j)}{n_j}.
    \label{eq:Hq2}
\end{align}
Only the two zones involved contribute to the change of $H_q$, so we have 
\begin{align}
    \Delta H_q &= - [Q(y_1) - \delta]\log \frac{Q(y_1) - \delta}{n_1 - 1} - [Q(y_2) + \delta]\log \frac{Q(y_2) + \delta}{n_2 + 1} \nonumber \\
    & \phantom{=} + Q(y_1) \log \frac{Q(y_1)}{n_1} + Q(y_2) \log \frac{Q(y_2)}{n_2}.
\end{align}
Based on Assumption B, $n_1$ and $n_2$ are large number and thus $q_1$ and $q_2$ are negligible compared to $Q(y_1)$ and $Q(y_2)$.
Furthermore, under Assumption C, $\delta$, as a value at the boundary between the two zones, can be approximated by the mean of $q_1$ and $q_2$ and is also deemed negligible. 
With these conditions, we can expand $\Delta H_q$ as 
\begin{align}
    \Delta H_q &= q_2 - q_1 + \delta(\log q_1 - \log q_2 + \frac{1}{n_1} + \frac{1}{n_2}) + O(\delta^2) + O(\frac{1}{n_1^2}) + O(\frac{1}{n_2^2})\\
    &\approx q_2-q_1+\delta \log \frac{q_1}{q_2}.
\end{align}
The sign of $\Delta H_q$ during the maximization of $H_Q$ depends on the values of $q_1$ and $q_2$. 

Therefore, we have proven that optimizing information transmission and input probability distribution modeling are not identical objectives for our IPU model, 
with no guarantee that an optimum can be achieved for both simultaneously.
In Appendix~\ref{appendix:toy}, we present a toy example where the optimal values for these two optimization problems can be analytically solved and are indeed different.
The two goals can still coincide in specific contexts as in the case of ICA \citep{Cardoso1997}.
However, this equivalence should not be assumed as a general rule.
The above derivation also applies to multivariate scenarios, as no assumptions about one-dimensionality of the input were made.

A pragmatical method to achieve the two goals is to find a suitable compromise, akin to the approach taken by sparse coding methods \citep{Olshausen1996}.
Minimizing the mean squared reconstruction error used in sparse coding is a reasonable first approximation of 
maximizing the mutual information between inputs and outputs \citep{vincent2003synaptic, baldi1995learning}.
It also approximates the minimization of the KL-divergence between the actual and modeled probability distributions \citep{olshausen1997sparse}.
The primary drawback of this approach is that neither of the two optimization goals is achieved optimally.

For early-stage IPUs, we argue that optimizing information transfer is more pressing and critical than accurately learning the input probability distribution.
Early-stage IPUs should prioritize maximizing $H_Q$ to produce an even output probability distribution, thereby retaining maximum information from the input. 
While not the optimal solution for approximating $p(x)$ with a fixed number of output levels, $N$, 
it represents a reasonable approximation through a step function, as depicted in Fig.~\ref{fig:px_qx}~(a).
If a more refined modeling of $p(x)$ is necessary, we can increase the output resolution $N$ of the IPU. 
This approach to balancing the two objectives in our discrete IPU model is referred to as even coding.

\section{Two Pixels}
For two-pixel input, $(x_a, x_b)$ or $\mathbf{x}$, in $X=\{\mathbf{x}_1, \mathbf{x}_2, \ldots, \mathbf{x}_M\}$, 
we have the option to either use a single IPU directly to process $\mathbf{x}$ or employ two IPUs to process $x_a$ and $x_b$ separately, 
followed by another IPU to process the combined outputs $(y_a, y_b)$. 
We will use the second approach, as processing as much information
locally reduces the cost of information transfer.
In fact, when images are stored on computers, gamma encoding is utilized 
to create an approximately even distribution of pixel values. 
When these images are displayed, pixel values undergo gamma correction to recover the original statistics for human eyes to process. 
In the following sections, we will assume that all pixel values $x$ have already been processed by dedicated IPUs, 
resulting in a roughly even probability distribution.

The probability distribution $p(x_a, x_a)$ of natural images is simple. 
The bulk of probability mass is concentrated around the diagonal line $x_a - x_b = 0$, with $p(x_a, x_b)$ rapidly decaying as $|x_a - x_b|$ increases
(refer to Fig.~\ref{fig:2px_partitions}~(a) as an example).

\subsection{One Output Dimension}
We first examine the scenario where there is only one output dimension, denoted as $y \in Y = \{y_1, y_2, \ldots, y_N\}$.
To investigate how IPUs learn $p(\mathbf{x})$, we conduct numerical experiments 
using a multilayer perceptron (MLP) as the IPU 
to approximate $y=f(\mathbf{x})$ and model $p(\mathbf{x})$ \citep{Montufar2014}.
MLP is very flexible and easy to train using contemporary deep learning frameworks such as PyTorch \citep{paszke2019pytorch}.
Other function approximation methods may also be applicable.

The chosen MLP architecture includes two inputs corresponding to the two pixel intensities and $N$ outputs corresponding to the $N$ output states. 
For each input, only one of the $N$ outputs should be activated.
To ensure the MLP is versatile enough to approximate complex functions, 
we use relatively large MLPs with two hidden layers of 200 and 100 nodes, respectively. 
The Rectified Linear Unit (ReLU) is used as the activation function for the two hidden layers.
We use the softmax function as the last layer of the MLP ensuring that each output value falls within the $(0, 1)$ range, with the sum of all outputs equating to 1. 

The IPU with one output dimension and $N$ output states is trained using the following loss function and stochastic gradient descent:
\begin{equation}
    E_{\text{OOD}} = \sum_{j=1}^N \langle y_{sj} \rangle_s \log \langle y_{sj} \rangle_s + k \langle -\sum_{j=1}^N y_{sj} \log y_{sj} \rangle_s.
\label{eq:loss_1d}
\end{equation}
$y_{sj}$ represents the value of the j-th output node for the s-th input sample, 
while $\langle \rangle_s$ denotes the average over all samples in a training batch. 

The softmax function ensures the outputs behave similarly to a probability distribution. 
The first term of the loss function guarantees that, on average, each output node has an equal opportunity to be activated, adhering to the principles of even coding. 
The second term fosters the activation of only one node per input sample while suppressing the others, thereby mimicking lateral inhibition. 
The coefficient $k$ is used to balance these two terms; we set $k=2/3$ for all experiments.

\begin{figure}
    \centering
    \begin{tabular}{ccc}
      \includegraphics[width=0.3\textwidth]{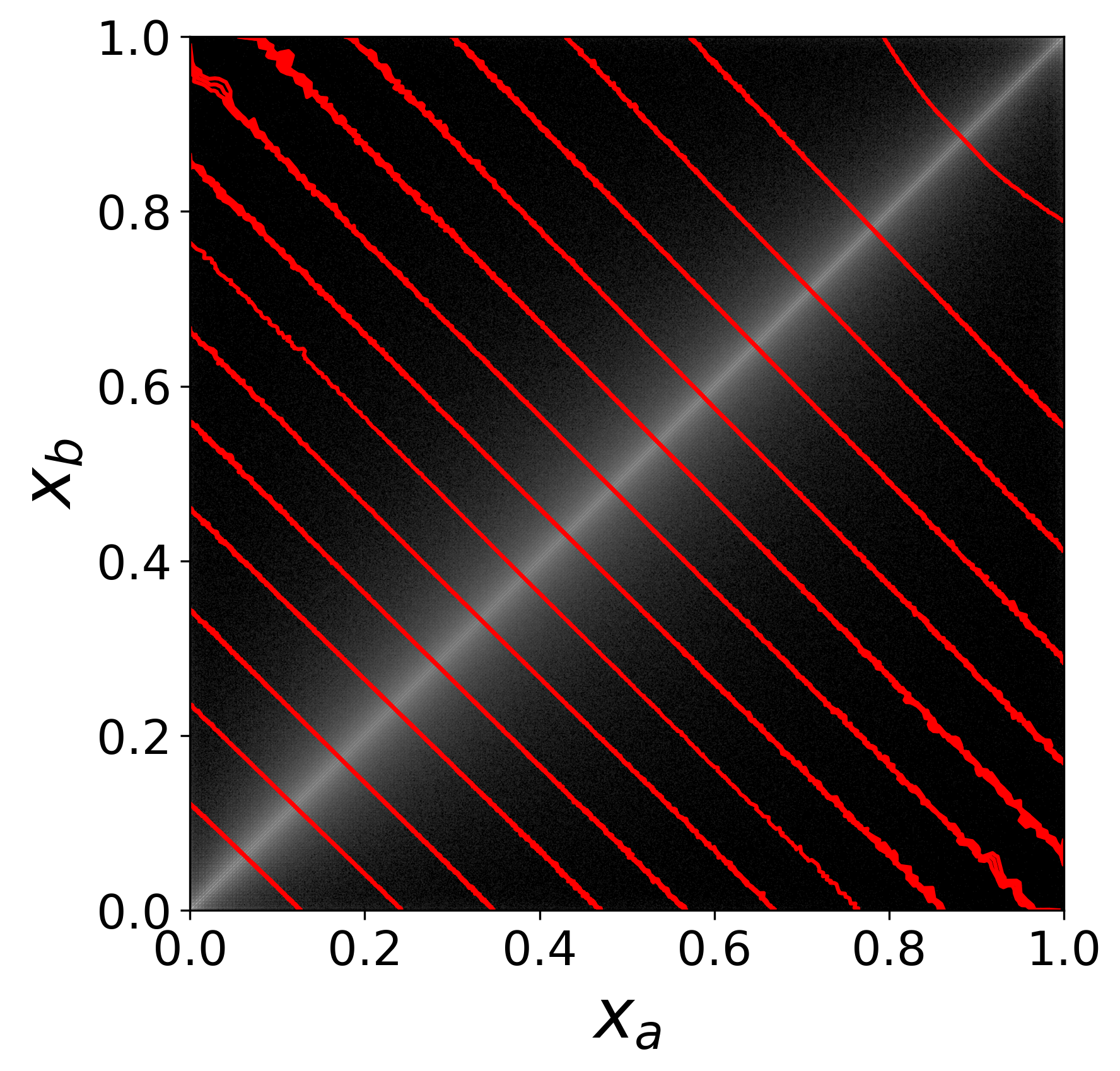} & \includegraphics[width=0.3\textwidth]{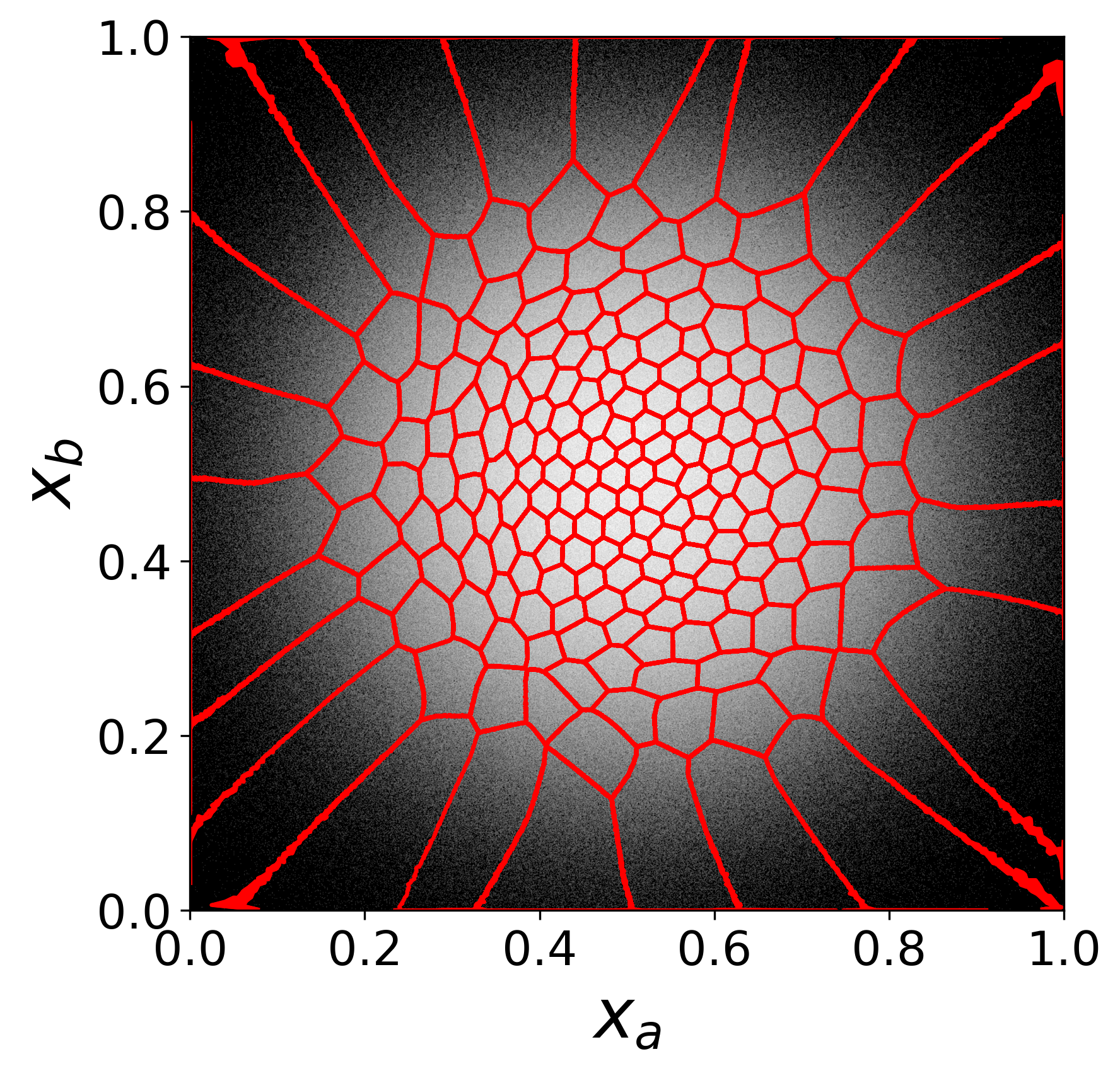} & \includegraphics[width=0.3\textwidth]{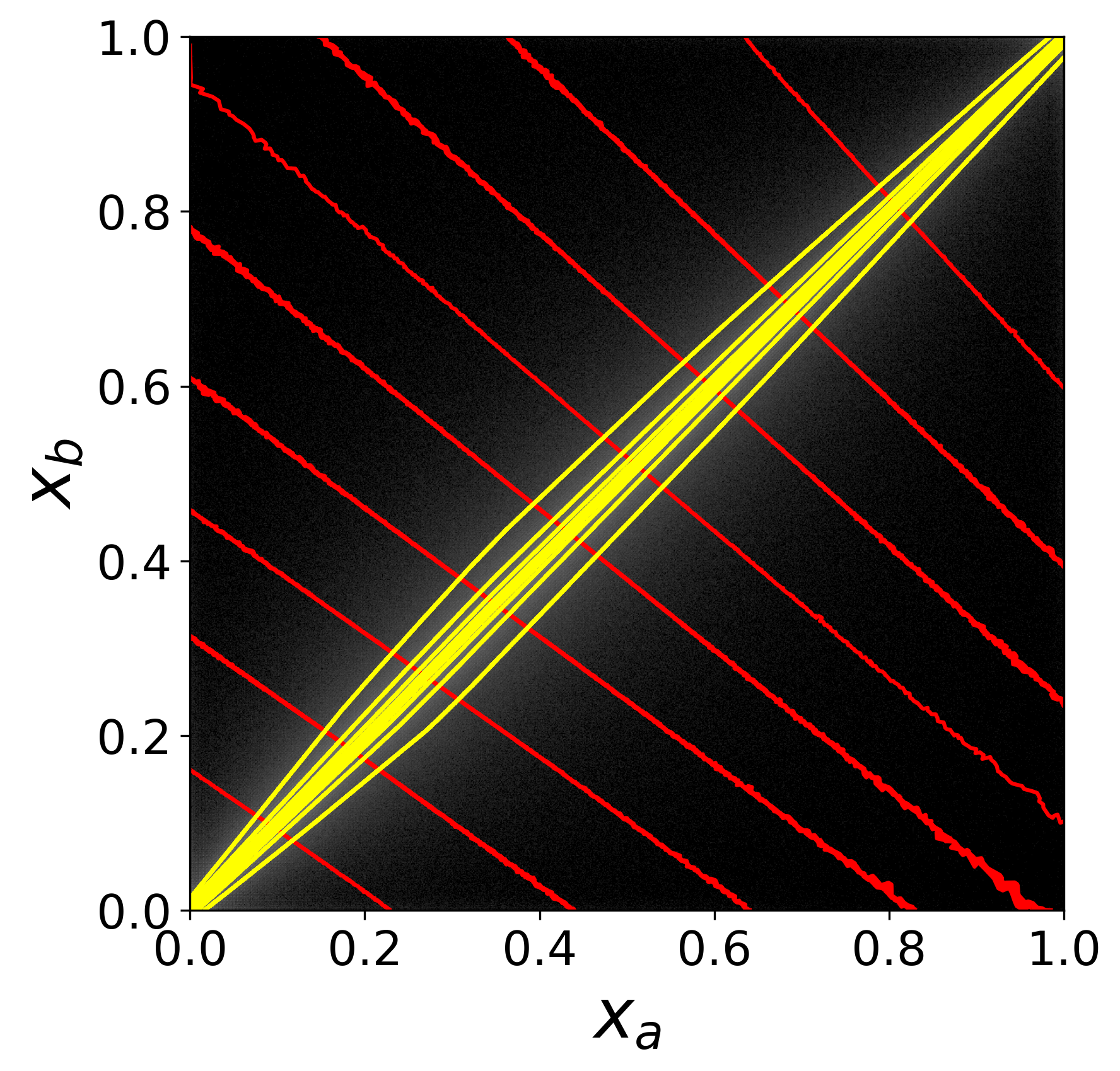} \\
    (a)  & (b) & (c) \\[6pt]
    \end{tabular}
    \caption{Evenly partitioning the two-pixel probability distribution learned by multilayer perceptrons (MLPs).
    The X and Y axes represent the intensities $x_a$ and $x_b$ of the two pixels. 
    The quantity $n(x_a, x_b) + 1$ is plotted in gray on a log scale, 
    where $n(x_a, x_b)$ denotes the number of occurrences of the two-pixel values among the sampled data. 
    Color lines indicate the boundaries of states for each output dimension learned by an MLP, with one color representing one dimension.
    (a) One output dimension with 16 states, which partitions the two pixel intensity space based on the total intensity $x_a + x_b$. 
    (b) One output dimension with 200 states partitions an artificial two pixel intensity which is a 2D normal distribution with a learned hexagonal lattice.
    (c) Two independent output dimensions, each with 10 states, 
        dividing the two pixel intensity space based on the total intensity $x_a + x_b$ and the contrast $x_a - x_b$ approximately.
    }
\label{fig:2px_partitions}
\end{figure}

For training data, we randomly sampled 10 million pairs of horizontally neighboring pixels from the COCO dataset \citep{COCO}. 
The models were trained for 20 epochs using Adam optimizer with a learning rate of 0.001 with a batch size 10000. 
Fig.~\ref{fig:2px_partitions}~(a) show the results learned by an MLP with 16 output nodes.
To demonstrate the flexibility of the method we also generated an artificial two pixel intensity probability distribution which is a 2D normal distribution as the training data.
Fig.~\ref{fig:2px_partitions}~(b) shows the results learned by an MLP with 200 output nodes. 
The size of the hexagonal lattice gradually increase from the center to the periphery to account the decrease of probability density 
so that each cell approximately contains same portion of probability.

\subsection{Multiple Independent Output Dimensions}
\label{section:md}
Intuitively, we can use two sets of lines parallel and perpendicular to $x_a - x_b = 0$ 
to divide the probability distribution into even partitions.
To achieve such partitioning, we need two independent output dimensions.
 Each dimension, $y_a$ and $y_b$, takes its value from the sets $Y_a$ and $Y_b$, respectively. 
 The output space, $Y$, is the Cartesian product $Y_a \times Y_b$, making the output $(y_a, y_b) \in Y$ a tuple. 
 If we denote the sizes of $Y_a$ and $Y_b$ as $N_a$ and $N_b$ respectively, then the total size of $Y$ is $N = N_a N_b$. 
 We utilize one IPU per output dimension, effectively combining them into a single IPU with $N$ output states.

 For independent outputs, also known as a factorial code, we have
\begin{equation}
    Q(y_a, y_b) = Q(y_a)Q(y_b).
\label{eq:two_feature_set_independent}
\end{equation}
Additionally, we aim for each output dimension to partition the input space evenly, leading to
\begin{equation}
    Q(y_a) = \frac{1}{N_a}, \quad Q(y_b) = \frac{1}{N_b}.
\label{eq:two_feature_set_even}
\end{equation}
By combining these two equations, we derive
\begin{equation}
    Q(y_a, y_b) = \frac{1}{N_a N_b}.
\label{eq:two_feature_set_combined_even}
\end{equation}
Conversely, from Eq.~(\ref{eq:two_feature_set_combined_even}), we can deduce Eq.~(\ref{eq:two_feature_set_independent}) and Eq.~(\ref{eq:two_feature_set_even}), 
indicating their equivalence. 
In other words, the maximization of output entropy leads to a factorial code and vice versa. 
This principle, also identified in other models \citep{Bell1995, Nadal2009}, may suggest its broader applicability as a general rule.

If more than two dimensions are required for partitioning the space, 
we can enforce Eq.~(\ref{eq:two_feature_set_combined_even})
for each combination of two dimensions to ensure independence between them. 
The loss function for multiple independent output dimensions is
\begin{equation}
    E_{\text{MIOD}} = \frac{1}{{D \choose 2}}\sum_{\langle d, d'\rangle} \sum_{jj'} \langle y_{dsj} y_{d'sj'} \rangle_s \log \langle y_{dsj} y_{d'sj'} \rangle_s 
    + \frac{k}{D} \langle -\sum_{d=1}^D \sum_j y_{dsj} \log y_{dsj} \rangle_s, 
\label{eq:loss_md}
\end{equation}
where $d$ is the dimension index, and $D$ is the number of dimensions. 
$\sum_{\langle d, d'\rangle}$ denotes the sum over all ${D \choose 2}$ combinations of two distinct dimensions.
$y_{dsj}$ represents the value of the j-th state in dimension $d$ for the s-th input sample.
The first term ensure all $N_a N_b$ combinations of $(y_a, y_b)$ are equally likely, satisfying Eq.~(\ref{eq:two_feature_set_combined_even}). 
The $E_{\text{MIOD}}$ loss function Eq.~(\ref{eq:loss_md}) is a generalization of Eq.~(\ref{eq:loss_1d}).

In our experiment, we use two MLPs to model the two IPUs respectively. 
They have the same size as the model used for one output dimension except each having 10 output states.
We follow the same training procedure as in the single MLP case.
Fig.~\ref{fig:2px_partitions}~(c) shows the result of partitioning using two independent output dimensions. 
Additional results are presented in Appendix~\ref{appendix:2px}.

Even though we introduce the loss functions Eq.~(\ref{eq:loss_1d}) and Eq.~(\ref{eq:loss_md}) with two pixel system as example,
They can be used beyond 2D and even to high dimensional cases such as small image patches. 
Additionally, the model with multiple independent output dimensions might model grid cells \citep{Hafting2005} 
in the entorhinal cortex, though this topic is beyond the scope of the current paper.

\section{Image Patches}
\label{section:image_patches}
Next, we move on to study image patches.
The multivariate input probability distribution $p(\mathbf{x})$ is considerably more complex compared to the previous examples \citep{simoncelli2001natural, hyvarinen2009natural, Lee2003}.
In principle, the loss functions formulated in the previous section can be adapted for image patches. 
While our preliminary results are promising, certain issues arise.

First, it is desirable for outputs to serve as meaningful representations, 
that is, they must have sufficient resolution to differentiate between distinct patches such as sky or flower, 
rather than grouping conceptually different patches together solely to achieve even distributions. 
In this scenario, one should not expect output states like $Q(\text{sky})$ and $Q(\text{flower})$ to have equal probabilities.

Second, we want the representation to directly reflect the similarity between image patches. 
Image similarity can be judged from different aspects and is thus inherently multi-dimensional. 
The model discussed in Section~\ref{section:md} is close to such a representation. 
However, a metric still needs to be defined. 
  
Third, the models discussed in the previous section perform well with simple inputs 
but are computationally expensive for complex inputs, such as image patches.
Using just one output dimension would require an impractically large number of output states, 
and ensuring independence across many output dimensions is also computationally intensive. 
Besides, determining the optimal number of dimension and the number of states for each dimension is an open question.  

\subsection{Model}

To address the first challenge above, we can relax the requirement for an even distribution at the most granular level while enforcing it on a larger scale in the transformed space. 
Consider the output distribution as analogous to the matter density distribution in a glass of water: uneven at the atomic level yet homogeneous on larger scales. 
This approach eliminates the need to require equal output probabilities for distinct outputs like $Q(\text{sky})$ and $Q(\text{flower})$. 
Instead, we can imagine a virtual coarse grid within the output space to partition the space, applying the even coding method at this macro level. 
Each cell in this grid may contain as few as one output state $y$ if $Q(y)$ is large, or multiple output states if each $Q(y)$ is small, 
ensuring that the total probability for each cell is approximately the same.

To solve the second problem, we propose the use of real-valued vectors $\mathbf{y} \in \mathbb{R}^D$ as outputs from the model, 
where $D$ represents the output dimension. 
Vectors within $\mathbb{R}^D$ possess well-defined metrics and have been utilized to represent a wide array of entities, 
including images, text, and categorical variables \citep{Hinton2006, Mikolov2013, guo2016entity}. 

Addressing the final challenge requires a new, efficient loss function that ensures an even distribution of outputs in the transformed space at a coarse level. 
Homogeneous distributions, often observed in nature, typically arise from equilibrium states. 
For example, electrical charges of the same sign repel and evenly distribute themselves across a spherical conductor's surface. 
Inspired by this natural phenomenon, we propose a loss function designed to make input samples repel each other in the transformed space
to promote an even distribution. This repulsive force decreases with distance, as described by
\begin{equation}
E =  \langle -\log \|\mathbf{y}_s - \mathbf{y}_{s'}\|_1 \rangle_{\langle s, s'\rangle}.
\label{eq:microscopic_loss_dense}
\end{equation}
Here, $\|\|_1$ denotes the $\ell ^{1}$ norm, and $\|\mathbf{y}_s - \mathbf{y}_{s'}\|_1$ is the  Manhattan distance 
between the D-dimensional representations of samples $s$ and $s'$.
The term $-\log \|\mathbf{y}_s - \mathbf{y}_{s'}\|_1$ represents the potential energy due to the repulsive force, 
which is inversely proportional to their Manhattan distance.
Alternative forms of potential energy and distance measures could also be applicable.
The notation $\langle \rangle_{\langle s, s'\rangle}$ denotes the average taken over all unique pairs of samples.

Should numerous samples converge at one point in the transformed space, they will exert a strong repulsive force in the surrounding area, 
thereby discouraging other samples from occupying nearby positions.
To prevent samples from pushing each other infinitely far apart, we restrict the representation values to be within the range $[0, 1]$.
With this constraint, the repulsive force pushes samples towards the vertices of the unit hypercube, 
effectively reducing the representations from real vectors to binary vectors as shown in Fig.~\ref{fig:outputs}~(a). 
As a result, an even distribution is achieved on a larger scale in the transformed space, which consists solely of the vertices.

With the exception of having a metric, this binary representation can be viewed as a special case of the representation discussed in Section~\ref{section:md}, where each dimension has only two states.
As previously demonstrated, an even distribution is equivalent to independent outputs. 
Therefore, we propose an alternative form of the loss function: rather than making samples repel each other in the output space, 
we aim to make the activation patterns of each node for the same sequence of samples repel one another. 
This encourages the output nodes to be as independent as possible. 
This is referred to as the node-wise loss function, in contrast to the original sample-wise loss function. 
The node-wise loss function is
\begin{equation}
    E =  \langle -\log \|\mathbf{y}_d - \mathbf{y}_{d'}\|_1 \rangle_{\langle d, d'\rangle},
\end{equation}
where $\mathbf{y}_d$ represents the activation pattern of node $d$ to a batch of training samples, and $\langle \rangle_{\langle d, d'\rangle}$ denotes the average taken over all unique pairs of nodes.

So far, the two binary states are symmetrical, resulting in the activation probability for each node being approximately 0.5.
However, an asymmetrical binary state configuration might be preferred in certain situations. 
For example, it can reflect the asymmetry in the energy required for a biological neuron to be in activated or inactive states \citep{attwell2001energy}.
Furthermore, with highly asymmetrical binary states, the representation becomes sparse. 
Previous studies have suggested that sparse representation offers numerous advantages \citep{Field1994, foldiak2003sparse, olshausen2004sparse, Beyeler2019} 
and may also relate to the inherent sparsity of image statistics \citep{Lee2003}. 
Therefore, we introduce a second term to the loss function to incorporate asymmetry. The revised sample-wise (and similarly for node-wise) loss function is defined as
\begin{equation}
    E = \langle -\log \|\mathbf{y}_s - \mathbf{y}_{s'}\|_1 \rangle_{\langle s, s'\rangle} + \alpha \langle \|\mathbf{y}_s\|_1 \rangle _s,
    \label{eq:microscopic_loss}
\end{equation}
where $\alpha$ is a free non-negative parameter to adjust asymmetry or sparsity. In our experiments below we use a $\alpha$ around 0.05. 
This effectively compresses the output states into a smaller subset of all vertices on the unit hypercube. 
Within this subspace, as we will demonstrate later, the first term in the loss function predominates, yielding an approximately even distribution.

In practice, we add a small value $\epsilon = 10^{-38}$ to the distance in the sample-wise loss function, allowing slightly different samples to share the same representation and enhancing numerical stability. 
Another approach for better numerical stability is to use the node-wise loss function. We find both approaches lead to qualitatively similar results.

\subsection{Experiments}
In the following experiments, we either use a single MLP with N outputs or N MLPs, each with one output,
as the IPU to approximate the transformation function $\mathbf{y} = f(\mathbf{x})$ and model $p(\mathbf{x})$.
The last layer of the MLP is a sigmoid layer, ensuring the output value ranges between 0 and 1. 
Our training data consist of random image patches extracted from the COCO 2017 image dataset \citep{COCO} or the ImageNet dataset \citep{deng2009imagenet}. 
Unlike ICA or sparse coding, we do not use any image preprocessing. Additional training details are provided in Appendix~\ref{appendix:imagepatch_exp}.

\subsubsection{Output Statistics}

\begin{figure}
    \centering   
    \begin{tabular}{cc}
        \includegraphics[width=0.45\textwidth]{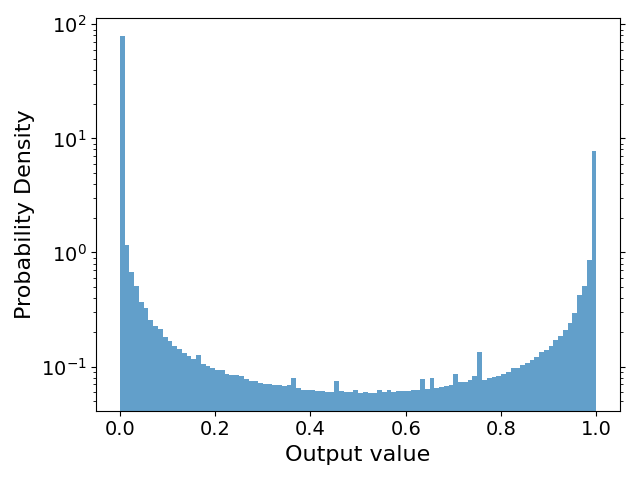} &   \includegraphics[width=0.45\textwidth]{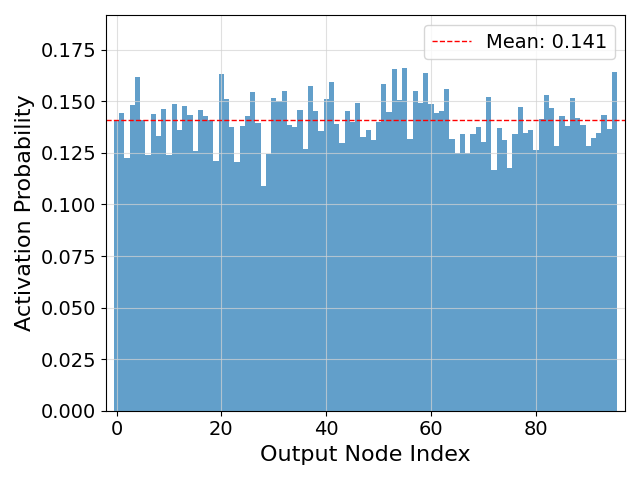} \\
      (a) & (b)
      \end{tabular}
      \caption{Statistical analysis of the learned representation using node-wise loss function with $\alpha = 0.625$. 
      (a) Histogram of the model's output values on a log scale. 
      The vast majority of the output values are either at 0 or 1, signifying that our model encoded the images using binary representation.
      (b) Probability of an output node being activated by a random image patch.
      }
\label{fig:outputs}
\end{figure}

First, we analyze the statistics of the learned representation. 
Across all experiments, we observe qualitatively similar output statistics, irrespective of the IPU architectures, 
variants of the loss function and training specifics, provided the training has properly converged. 
For illustration, we present an example using an IPU model trained on $5 \times 5$ color image patches with node-wise loss function. 
It uses 96 MLPs, each with one output node and a middle layer of 48 nodes.
Following training, the model is used to generate representations for 1 million random image patches for this analysis.

Fig.~\ref{fig:outputs}~(a) presents the histogram of output values on a logarithmic scale. 
Though the outputs can be any real value between 0 and 1, 
after the model has been trained the vast majority of the output values are either at 0 or 1, signifying that our model encoded the images using binary representation.
As such, after training, we can round the outputs to yield a true binary representation.

Fig.~\ref{fig:outputs}~(b) illustrates the probability of an output node being activated by a random image patch. 
All nodes have similar activation probabilities, indicating an even distribution at the coarse scale across all nodes. 
Further statistical analysis of the output representation is in Appendix~\ref{appendix:output_stat} 
confirms that even distribution is also achieved at smaller scales.
Therefore, the nonlinear IPU model can achieve even output distribution and statistically independent responses in a single step, 
eliminating the need for preprocessing steps like whitening or postprocessing steps such as divisive normalization \citep{schwartz2001natural}. 

\subsubsection{Image Patch Similarity}
\begin{figure}
    \centering
    \begin{tabular}{cc}
        \includegraphics[width=0.3\textwidth]{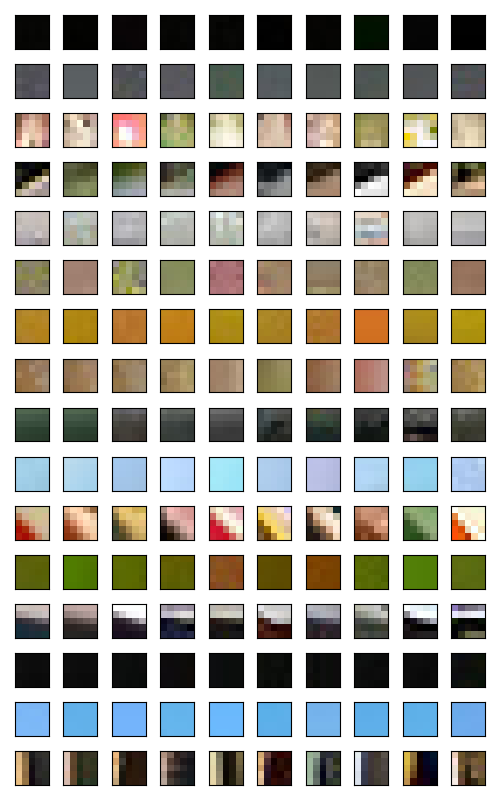} & \includegraphics[width=0.3\textwidth]{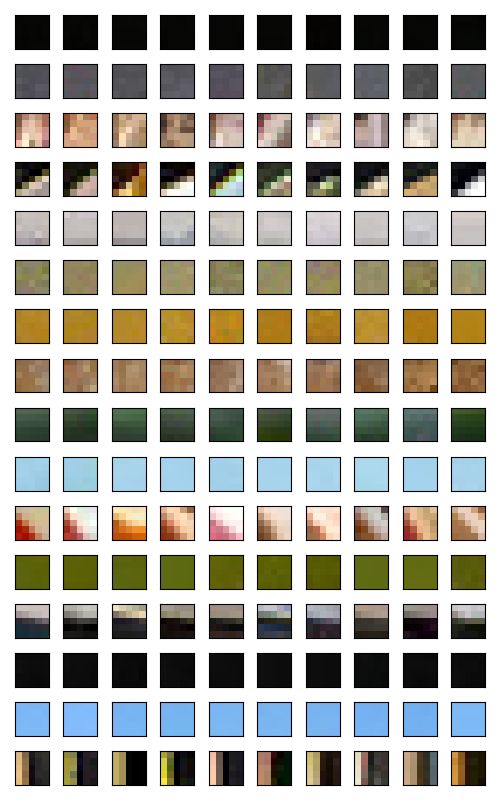} \\
      (a) Even coding model with 96 binary outputs & (b) Deep learning model with 128 float outputs
      \end{tabular}
      \caption{Image patches with the shortest distance in the representation space to 16 randomly selected image patches. 
      The first column displays the 16 random image patches, while the succeeding nine columns display patches that are closest to the first-column patches in the same row.
      (a) Distances are computed using an even coding IPU model, trained through unsupervised learning, with 96 binary number outputs.
      (b) Distances are computed using the first 10 layers of a convolutional neural network (VGG16) model, pretrained through supervised learning, with 128 floating-point number outputs.
      }
\label{fig:similar_patches}
\end{figure}

Next, to examine how the learned representation reflects the similarities between image patches, 
we analyze image similarity in Fig.~\ref{fig:similar_patches}. To create this visualization, we use the following steps:
We randomly sample 1 million $5 \times 5$ color image patches from the datasets.
For each image patch, we use the same $5 \times 5$ color image patch model as before to generate a representation, which is a binary vector of size 96 (12 bytes).
For comparison with deep learning methods, we also use the first 10 layers of a convolutional neural network (VGG16) \citep{simonyan2014very} pre-trained on ImageNet through supervised learning
to generate another representation for each image patch, which is a floating-point vector of size 128 (512 bytes).
Out of the 1 million, we selected 16 random image patches and plotted them as the first column in Fig.~\ref{fig:similar_patches}~(a) and (b).
For each of the 16 random images, we calculated the distance to the remaining images using the representations generated before, 
and we chose the top 9 image patches with the smallest distances and showed them in the remaining 9 columns in Fig.~\ref{fig:similar_patches}~(a) and (b) in order, respectively. 
   
We can see in Fig.~\ref{fig:similar_patches}~(a) the learned representation clearly captures perceptual similarity. 
The results shown in Fig.~\ref{fig:outputs}, Fig.~\ref{fig:similar_patches}~(a),
and additional results in Appendix~\ref{appendix:output_stat} confirm that we can indeed learn a meaningful representation which reflects the image similarity while adhering to even coding.
Also, our IPU model produces comparable results using unsupervised training to those generated with the deep learning model using supervised training while using less 
than 3\% of the memory for the representations.

\subsubsection{Local Edge Detectors and Orientation-Selective Units}
\begin{figure}
    \centering
        \includegraphics[width=\textwidth]{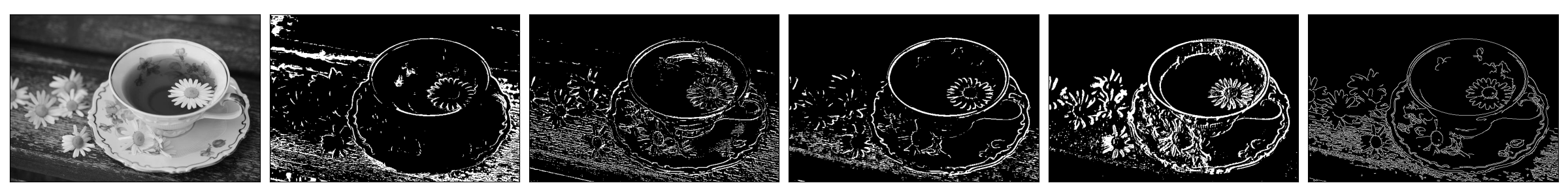} \\
        \includegraphics[width=\textwidth]{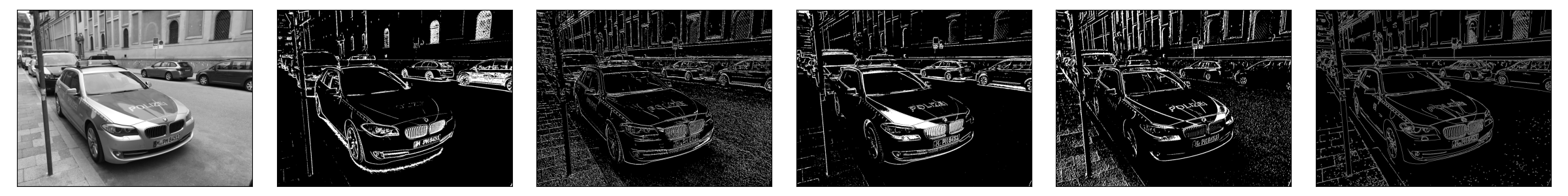} \\
      \caption{
        Feature maps of nodes resembling local edge detectors. 
        The first column presents the grayscale test images. 
        Each subsequent column, except the last one, displays the feature maps corresponding to the same output node for the test images. 
        The last column shows edges generated by the multi-stage Canny edge detector for comparison.
      }
\label{fig:edge}
\end{figure}

\begin{figure}
    \centering
    \includegraphics[width=\textwidth]{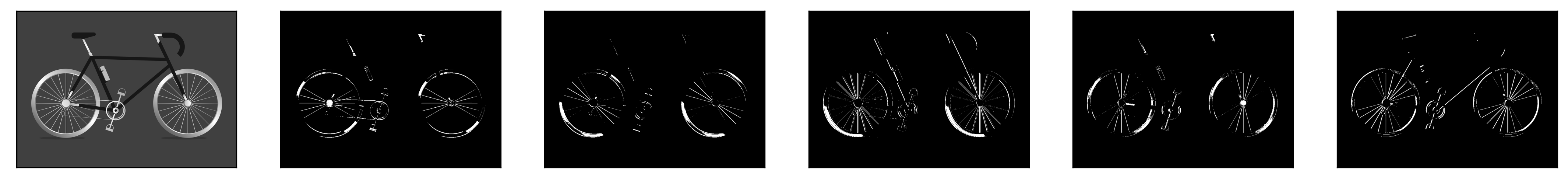}
    \caption{Test image and feature maps of 5 orientation-selective nodes. 
      }
\label{fig:orientation}
\end{figure}

Early visual systems possess local edge detectors and orientation-selective units \citep{levick1967receptive, baden2016functional}. 
It has been proposed that these features are the results of efficient coding \citep{barlow1989unsupervised, Atick1992}, 
and this has been supported by methods based on information theoretical principles, such as ICA \citep{bell1997independent} and sparse coding \citep{Olshausen1996}.
Although convolutional neural networks have achieved tremendous success in computer vision 
and are frequently compared to the biological visual system for their hierarchical processing of information \citep{lecun2015deep},
they are not based on efficient coding principles, and their initial layers have not shown proficiency in edge detection \citep{le2021revisiting}. 
Notably, prevalent local edge detection algorithms, such as the Canny edge detector \citep{canny1986computational}, still rely on non-deep learning methods.

The even coding IPU model is based on information theoretical principles, does it have local edge detectors and orientation-selective units as one would expect?
To answer this, we trained an IPU on $4 \times 4$ grayscale image patches and applied it to images with a stride of 1 pixel, generating feature maps for each output node. 
The model comprises a MLP with 64 outputs and an intermediary layer with 100 nodes. 
Fig.~\ref{fig:edge} illustrates the feature maps of 4 output nodes of the IPU.
It also shows edges generated by the Canny edge detector as comparison. 
Interestingly, with this simple network architecture, the even coding IPU model demonstrated a remarkable capability in edge detection.
Furthermore, Fig.~\ref{fig:orientation} shows the feature maps of 5 output nodes for a sample bike image.
Spokes of different orientations activate different nodes, indicating that these output nodes have varying orientation preferences, 
similar to orientation-selective units found in biological systems.

\subsubsection{Luminance and Color}
Previous studies on image patch representation have often prioritized contrast while neglecting the average luminance. 
For example, in methods such as ICA or sparse coding, average luminance information is typically lost during preprocessing steps. 
However, omitting luminance prevents the development of a comprehensive model for image patches and limits our understanding of the early visual system. 
Firstly, a significant portion of image patches, such as those extracted from skies or skin, exhibit minimal contrast and are characterized predominantly by their luminance and color. 
Secondly, the encoding of luminance and color information by the visual system, along with its interaction with other properties, remains a critical area of research
\citep{barlow1978intensity, milner2017population, vinke2020luminance, Rossi1996, mante2005independence, li2015mixing, wienbar2018dynamic}.

\begin{figure}
    \centering   
    \begin{tabular}{cc}
        \includegraphics[width=0.45\textwidth]{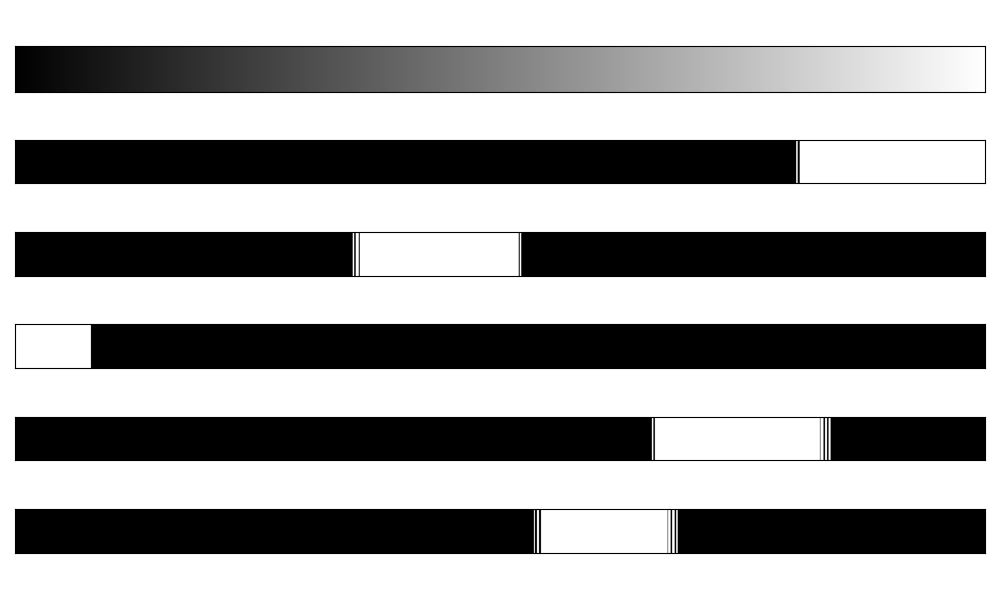} &   \includegraphics[width=0.45\textwidth]{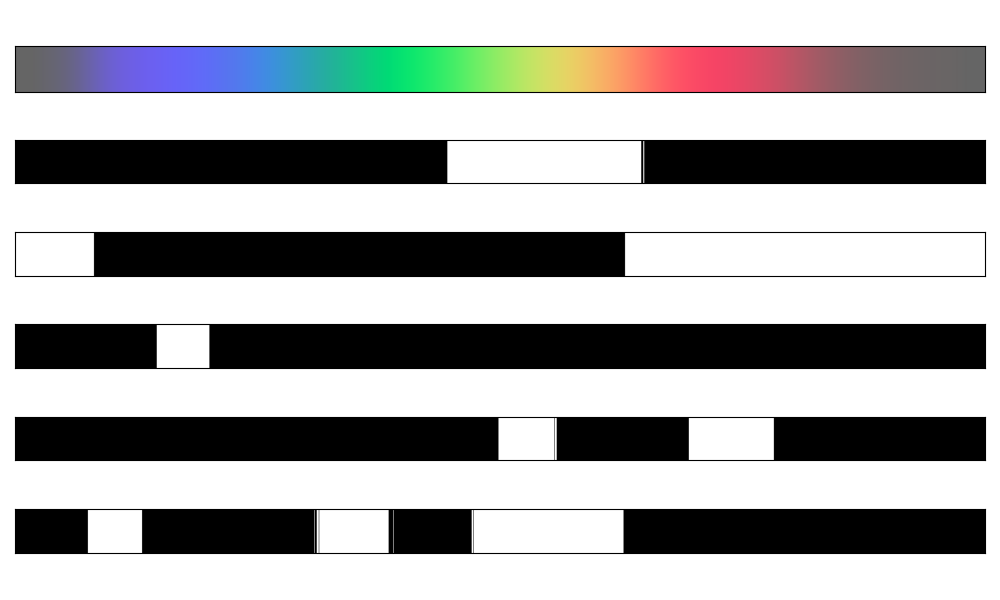} \\
      (a) & (b)
      \end{tabular}
      \caption{IPU's response to varying gray luminance levels (a) and spectral colors (b). The first row displays the input images.
      Each subsequent row shows the response of the same node to both types of input.
      }
\label{fig:color_luminance}
\end{figure}

When training even coding IPU model, we refrain from using any image preprocessing techniques, such as whitening, to preserve luminance information.
As a result, the image IPU must learn to encode varying luminance levels in addition to contrast, using a binary population code.
To explore the IPU's response to different homogeneous (approximately) luminance levels,
we applied the same $5 \times 5$ color IPU  to an input image with varying luminance on a linear grayscale, as shown in Fig.~\ref{fig:color_luminance}~(a).
Approximately 60\% of nodes are activated by the gray image.
We plot the input image and feature maps of 5 nodes in Fig.~\ref{fig:color_luminance}~(a) 
(refer to Appendix~\ref{appendix:color_luminance_all} for more information).
These results are consistent with experimental findings showing that neurons in the early visual system exhibit 
varying luminance tuning ranges \citep{hammon1982luminance, volgyi2004convergence}.
Additionally, they align with research indicating that light intensity is collectively encoded by neurons 
\citep{milner2017population}.

We also tested the IPU model on an image showing the visible spectrum on a linear scale \citep{wikimedia_visible_spectrum}
to understand its response to different homogeneous (approximately) colors.
The responses from some color-activated nodes are displayed in Fig.~\ref{fig:color_luminance}~(b). 
We used the same nodes in the same order as in Fig.~\ref{fig:color_luminance}~(a).
The first and second nodes are activated and inhibited, respectively, by a single, broad segment of colors on the spectrum. 
The third node shows more specific color selectivity and is only activated by a narrow segment of the color spectrum. 
Unlike the case with luminance, where nodes are activated only by a single segment on the grayscale, 
some nodes like the fourth and fifth are tuned to multiple segments of colors on the spectrum.
These results are consistent with experimental findings showing that neurons have different chromatic tuning properties 
to encode color using a distributed code \citep{wachtler2003representation}. 
Overall, we found that approximately three-quarters of the output nodes could be activated by some portion of the spectrum input 
(see Appendix~\ref{appendix:color_luminance_all}).
This aligns with experimental data suggesting that about 60\% to 80\% of V1 neurons are color-selective \citep{johnson2001spatial, friedman2003coding, thorell1984spatial, hass2013v1}. 
More specifically, about one-quarter of nodes are only activated/inhibited by a single spectrum segment, 
and about half of the nodes are activated by multiple spectrum segments. 
No experimental evidence has yet been found to support or contradict these numbers.

The results also highlight that the same node can exhibit selectivity for different stimulus attributes such as luminance, color, edge, and orientation. 
This is consistent with the finding that neurons in the early visual system multiplex information about multiple stimulus properties 
\citep{Rossi1996, friedman2003coding}.

\section{Discussion}
\subsection{Related Work}
\textbf{Efficient sensory coding.}
How sensory neurons encode information efficiently using population code has been examined via spiking neuron models \citep{ganguli2014efficient, yerxa2020efficient}.
These studies, to make the problem tractable, employ numerous assumptions and approximations, 
such as rate coding, Poisson spike generation, specific neuron tuning curve forms, and Fisher information as the lower bound of mutual information. 
While these assumptions are reasonable and widely used, they impose certain limitations on the applicability of these studies. 
In contrast, our ab initio approach aims to distill the essence of an information processing system with just four basic assumptions, 
rendering our model both simpler to study and more general.

Our model, despite its simplicity, yields results comparable to those from spiking neuron models.
For example, the output states in our model correspond to neurons in Ganguli and Simoncelli's model \citep{ganguli2014efficient},
where each neuron is tuned to a specific portion of the input stimuli, as dictated by its tuning curve. 
Similarly, in our IPU model, each output is only responsive to a specific portion of the input, defined by the step function $y=f(x)$.
Each output state of IPU corresponds to an equal portion of $p(x)$ means the distribution of IPU output states is proportional to $p(x)$ in the continuous limit.
Thus our Eq.~(\ref{eq:qx2}) is analogous to their Eq.~(2.14).
Furthermore, the concept that the input probability distribution $p(x)$ is implicitly encoded in the transformation $y=f(x)$ mirrors their findings.

Another example of similarity can be seen in the two-pixel case, as shown in Fig.~\ref{fig:2px_partitions}~(b), which resembles the 2D case study presented in Fig.~3 by Yerxa et al. \citep{yerxa2020efficient}.
In our model, the hexagonal lattice structure is learned, contrasting with their study, where it is predefined.

Therefore, our IPU model offers a simplified, complementary approach for studying sensory neuron information encoding alongside spiking neuron models.

\textbf{Independent component analysis.}
Our method shares many similarities with Bell and Sejnowski's infomax ICA algorithm \citep{Bell1995, bell1997independent}.
Both extend Laughlin's work \citep{Laughlin1981} to multiple dimensions. 
Both methods maximize output entropy of a zero-noise system to achieve maximal information transmission, 
and can be considered as a generalisation of Linsker's infomax principle \citep{Linsker1988}.

The main differences between ICA and our approach is the assumptions we take.
ICA assumes inputs are linear superposition of independent sources while our method does not have the "sources" concept.
ICA also assumes input and output to be continuous variables and the mapping between them are of the form 
$\mathbf{y} = g(\mathbf{Wx} + \mathbf{x}_0)$,
where $W$ is the square matrix and g is an invertible function. 
Our method assumes the input and output are discrete and the mapping from input to output 
is a many-to-one step function as defined by Eq.~(\ref{eq:y}).
Therefore, the choice of transformation of our method is much more flexible,
and it also does not require the output and input dimensions to have the same size.

Our method is not suitable for problems ICA is intended to solve such as blind separation of continuous signals.
Only in case overcoming the limitation of ICA's assumptions is a major concern and computational resource is not a limiting factor,
one might use Eq.~(\ref{eq:loss_md}) with a large number of states for each dimension to approximate continuous output 
as a potential alternative to ICA. However, this is out of the scope of this paper.

\textbf{Sparse coding.}
Both sparse coding \citep{Olshausen1996, olshausen1997sparse} and even coding yield edge and orientation selective nodes. 
Our loss function for image patches, Eq.~(\ref{eq:microscopic_loss}), also incorporates a sparsity term. 
The primary differences between our method and sparse coding are:

1. Sparse coding, similar to ICA, assumes inputs as linear superpositions of basis functions, limiting the type of features it can learn. 
Our method, devoid of such assumptions, expresses features as a general function of the inputs.

2. Sparse coding use image reconstruction error minimization as the optimization goal. 
However, this is merely a reasonable first approximation of rate of transmission which assumes images are Gaussian distributed \citep{vincent2003synaptic}.
Our method directly maximizes the rate of transmission.

3. Calculating image reconstruction error requires knowing both input and output values and deriving the input from the outputs.
These requirements may limit the biological plausibility of the method.
Conversely, even coding, which solely requires local knowledge of the output, offers a more biologically plausible model for neural implementation.

4. Our model identifies luminance and color selective nodes and can be readily applied to video inputs.

\subsection{Conclusion}

In summary, we abstract the complex biological early visual system using four assumptions that serve as the foundation for all studies in this paper. 
We prove that maximizing information transmission and modeling the input probability distribution are not identical objectives. 
However, early-stage IPUs can pragmatically pursue both objectives using our even coding method.

According to even coding, we propose two types of unsupervised loss functions. 
The first explicitly enforces output statistics using 2D systems as examples, but it also applies to higher dimensions. 
The second achieves even coding implicitly by making response vectors repel each other, thereby resulting in a binary representation, 
as demonstrated in the image patch cases. 
Both types of loss functions require only local knowledge at the outputs, 
providing a more biologically plausible model for neural implementation compared to works that require non-local information for both input and output.

We confirm that the trained image patch IPU model produces an approximately even output distribution and statistically independent responses. 
It also exhibits remarkable similarities to early visual systems, including multiplexing, nonlinear population code, 
binary signals, and independent outputs \citep{ecker2010decorrelated}. 
Additionally, it features local edge-detecting and orientation-selective units, as well as nodes with different luminance and chromatic tuning properties. 
Moreover, when compared to a deep learning model, our image patch IPU model significantly outperforms in efficiency with only unsupervised learning.

There are several intriguing directions for future research:
\begin{enumerate}
    \item The IPU model applies to a wide range of inputs, from simple one-pixel instances to complex color image patches. 
    Could the model's application extend beyond the early stages of visual information processing? 
    For inspiration, preliminary experiments with multiple layers of the IPU model trained on larger image patches have not yet revealed the emergence of object-specific features. 
    However, given that these experiments were conducted with limited computational resources (using only one Nvidia GTX 1660 Super GPU), the findings are not definitive.
    
    \item The even coding model could be readily extended to process video data by adding a time dimension. The outputs from the nodes resemble temporal raster plot. 
    A detailed comparison with early visual systems could provide significant insights.
    
    \item The even coding model also has the potential to adapt to binocular vision data by incorporating an additional input dimension of size two. 
    Investigating whether the model can detect binocular disparity or even construct a 3D model of the world would be fascinating.

    \item While this paper primarily addresses visual information, the versatility of the even coding model suggests its applicability to modeling other multivariate probability distributions.
    
    \item An important area for future research is to explore how biological systems might implement even coding. 
    Are mechanisms such as lateral inhibition and homeostatic plasticity involved?

    \item Lastly, a crucial question remains: how can the implicitly encoded prior probability distribution be leveraged for inference?
\end{enumerate}

\begin{ack}
I am grateful for the interdisciplinary doctoral fellowship (IDK-NBT) provided by the Center for NanoScience (CeNS) of LMU 
and the Elite Network of Bavaria, which funded my attendance at NeurIPS 2011, where I became inspired to work on this problem. 
I would like to thank Jonathan Shock, Dean Rance, Jonathan Pillow and E.J. Chichilnisky for their valuable comments and feedback 
on the manuscript of this paper.
I thank the anonymous reviewers of a previous version of this manuscript for their constructive comments, which greatly improved the presentation.
This research was conducted independently by the author, without the use of any institutional resources or the endorsement of any institution.
\end{ack}

\bibliography{main}

\appendix
\section{Proof: Maximizing Transmission Rate is Equivalent to Maximizing $H_Q$}
\label{appendix:rot}
The rate of transmission is defined according to Shannon's information theory \citep{Shannon1948} as: 
\begin{eqnarray}
    R &=& H_Q - H(y|x),
\end{eqnarray}
where $H(y|x)$ is the conditional entropy of $y$ given $x$. This can be defined as:
\begin{eqnarray}
    H(y|x) &=& -\sum_{i, j} p(x_i, y_j) \log p(y_j|x_i) \\
      &=& -\sum_{i, j} p(x_i) p(y_j|x_i) \log p(y_j|x_i) \\
      &=& -\sum_i \Big[p(x_i) \sum_j p(y_j|x_i) \log p(y_j|x_i)\Big],
\end{eqnarray}
where $p(x_i, y_j)$ is the joint probability of $x$ and $y$ and $p(y_j|x_i)$ is the conditional probability - 
that is, the probability of the output value being $y_j$ given that we know the input is $x_i$.

Now, for a deterministic model where each input $x_i$ maps to one and only one output $y_j$, we can have $p(y_j|x_i)$ being 1 for some $j$ and 0 for the others. 
As a result, the expression $p(y_j|x_i) \log p(y_j|x_i)$ is always 0, whether $p(y_j|x_i)$ is 0 or 1. 
Therefore, we have $H(y|x) = 0$.
Consequently, the rate of transmission $R$ simplifies to $R = H_Q$. 
Hence, maximizing the rate of transmission is equivalent to maximizing $H_Q$. 

\section{Proof: Cross-Entropy $H_{pq}$ is Equivalent to $H_q$}
\label{appendix:cross}
Under the condition Eq.~(\ref{eq:Qy}) and Eq.~(\ref{eq:qx2}) specified in the main text,
we can rewrite the cross-entropy $H_{pq}$ as:
\begin{eqnarray}
    H_{pq}
    &=& -\sum_{x} p(x) \log q(x) \\
    &=& - \sum_{j=1}^N \sum_{x \in G_j} p(x) \log q(x)\\
    &=& - \sum_{j=1}^N \sum_{x \in G_j} p(x) \log \frac{Q(y_j)}{n_j}\\
    &=& - \sum_{j=1}^N \log \frac{Q(y_j)}{n_j} \sum_{x \in G_j} p(x) \\
    &=& - \sum_{j=1}^N Q(y_j) \log \frac{Q(y_j)}{n_j}.
\end{eqnarray}
Similarly, we can express $H_q$ as:
\begin{eqnarray}
    H_q &=& -\sum_x q(x) \log q(x) \\
    &=& -\sum_{j=1}^N \sum_{x \in G_j} q(x) \log q(x) \\
    &=& -\sum_{j=1}^N \sum_{x \in G_j} q_j \log q_j \\
    &=&  - \sum_{j=1}^N Q(y_j) \log \frac{Q(y_j)}{n_j}.
\end{eqnarray}
Comparing the two results, we have 
\begin{equation}
    H_{pq} = H_q.
\end{equation}

\section{An example to show that optimizing transmission does not necessarily lead to optimal input probability modeling}
\label{appendix:toy}

\begin{figure}
    \centering
    \includegraphics[width=0.7\textwidth]{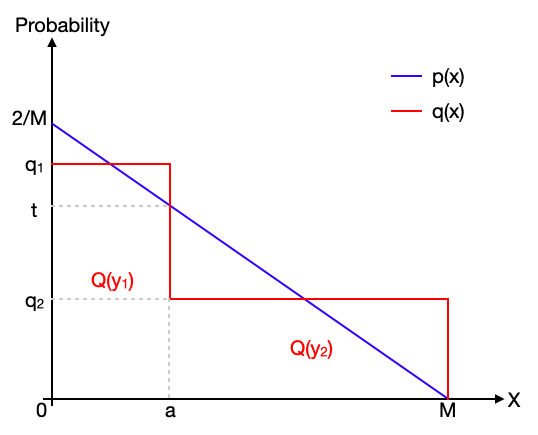}
    \caption{A toy example demonstrating the non-equivalence of the two goals. The input $x$ has M states. 
    The input distribution $p(x)$ decays linearly with $x$. 
    The IPU model has only two output states, with the first $a$ input states mapped to $y_1$ 
    and the remaining $M-a$ input states mapped to $y_2$. 
    Maximizing information transmission leads to $a = (1-\frac{1}{\sqrt{2}})M \approx 0.293M$. 
    Maximizing input distribution modeling leads to $a \approx 0.602M$.}
\label{fig:toy}
\end{figure}

In Section \ref{section:two_goals} of the main text, we demonstrate that optimizing information transmission is not the same as 
optimizing input distribution modeling for our IPU model. 
Now, we use a toy example to illustrate that the solutions to these two optimization problems are indeed different.

In this example, the input probability distribution $p(x)$ decays linearly, as shown in Fig.~\ref{fig:toy}. 
$M$ is a very large integer so that $p(x)$ is nearly indistinguishable from a continuous line in the plot. 
The highest probability of an $x$ state is $\frac{2}{M}$, ensuring the total probability sums to 1.

The IPU model in this example has only two output states and models $p(x)$ with an approximated probability distribution $q(x)$, 
represented by the step-shaped red line in Fig.~\ref{fig:toy}. 
The first $a$ input states in $G_1$ are mapped to the first output state $y_1$, 
and the rest $M-a$ input states in $G_2$ are mapped to the second output state $y_2$. 
$q(x)$ for all $x \in G_1$ equal to $q_1$, and for all $x \in G_2$ equal to $q_2$.

Drawing from Eq.~(\ref{eq:Qy}) and Eq.~(\ref{eq:px_qx}) in the main text, along with the formulas for the areas of a trapezoid and triangle, we derive the following relationships for the two output probabilities:
\begin{align}
    Q(y_1) &= a q_1  = \frac{a}{2}(t + \frac{2}{M}),
\label{eq:Q_y1}
\end{align}
and
\begin{align}
    Q(y_2) &= (M-a) q_2 = \frac{M-a}{2} t,
\label{eq:Q_y2}
\end{align}
where $t$ is an auxiliary quantity shown in Fig.~\ref{fig:toy}. 
Given that corresponding sides of similar triangles in Fig.~\ref{fig:toy} are proportional, we obtain the relation
\begin{equation}
    \frac{t}{2/M} = \frac{M-a}{M},
\end{equation}
leading to
\begin{equation}
    t = \frac{2(M-a)}{M^2}.
\label{eq:t}
\end{equation}

From Eq.~(\ref{eq:Q_y1}), Eq.~(\ref{eq:Q_y2}) and Eq.~(\ref{eq:t}) 
we can express $q_1$, $q_2$, $Q(y_1)$, and $Q(y_2)$ as 
\begin{align}
    q_1 &= \frac{2M-a}{M^2} \label{eq:q1} \\
    q_2 &= \frac{M-a}{M^2}  \label{eq:q2} \\
    Q(y_1) &= \frac{a(2M-a)}{M^2} \\
    Q(y_2) &= \frac{(M-a)^2}{M^2}.
\label{eq:q1_q2}
\end{align}

To optimize information transmission, we should have $Q(y_1) = Q(y_2) = \frac{1}{2}$. 
The solution is
\begin{equation}
    a = (1-\frac{1}{\sqrt{2}})M \approx 0.293M.
\label{eq:a_optim1}
\end{equation}

On the other hand, from Eq.~(\ref{eq:KL-divergence}) in main text, to optimize input
probability distribution modeling is to minimize $H_q$, which in the toy example is
\begin{align}
    H_q &= -\sum_x q(x) \log q(x) \\
        &= -\sum_j \sum_{x \in G_j} q_j \log q_j \\
        &= -a q_1 \log q_1 - (M-a) q_2 \log q_2.
\end{align}

\begin{figure}
    \centering
    \includegraphics[width=0.7\textwidth]{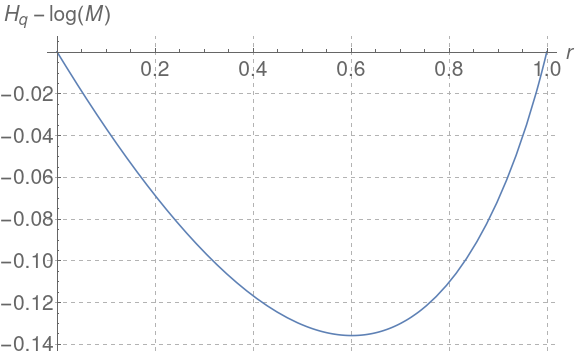}
    \caption{$H_q - \log M$ as a function of $r \equiv a/M$. $H_q$ reaches minimum at $a \approx 0.602M$.}
\label{fig:min_hq}
\end{figure}

Substitute $q_1$ and $q_2$ with Eq.~(\ref{eq:q1}) Eq.~(\ref{eq:q2}) we have
\begin{align}
    H_q &= -\frac{a (2M - a)}{M^2} \log \left(\frac{2 M - a}{M^2}\right) -  \frac{(M - a)^2}{M^2} \log \left(\frac{M-a}{M^2}\right).
\end{align}
Defining $r \equiv a/M$, this simplifies to
\begin{align}
    H_q &= -r (2-r) \log (2-r) - (1-r)^2 \log (1-r) + \log M.
\end{align}
Since $M$ is constant, we focus on minimizing $H_q - \log M$ instead. 
Fig.~\ref{fig:min_hq} plots $H_q - \log M$ for $r$ between 0 and 1, 
with numerical minimization yielding $r \approx 0.602$ or $a \approx 0.602M$. 
This solution is different with Eq.~(\ref{eq:a_optim1}). 
Therefore, we have demonstrated that the two optimization problems are distinct.

Additionally, this example shows that although we only aim to minimize information transformation, 
by increasing the number of output levels, $q(x)$ effectively approaches $p(x)$ as well. 
This is how we achieve both goals with the discrete even coding method.

\section{Supplementary Results for the Two-Pixel Case}
\label{appendix:2px}
\begin{figure}
    \centering
    \begin{tabular}{cc}
        \includegraphics[width=0.45\textwidth]{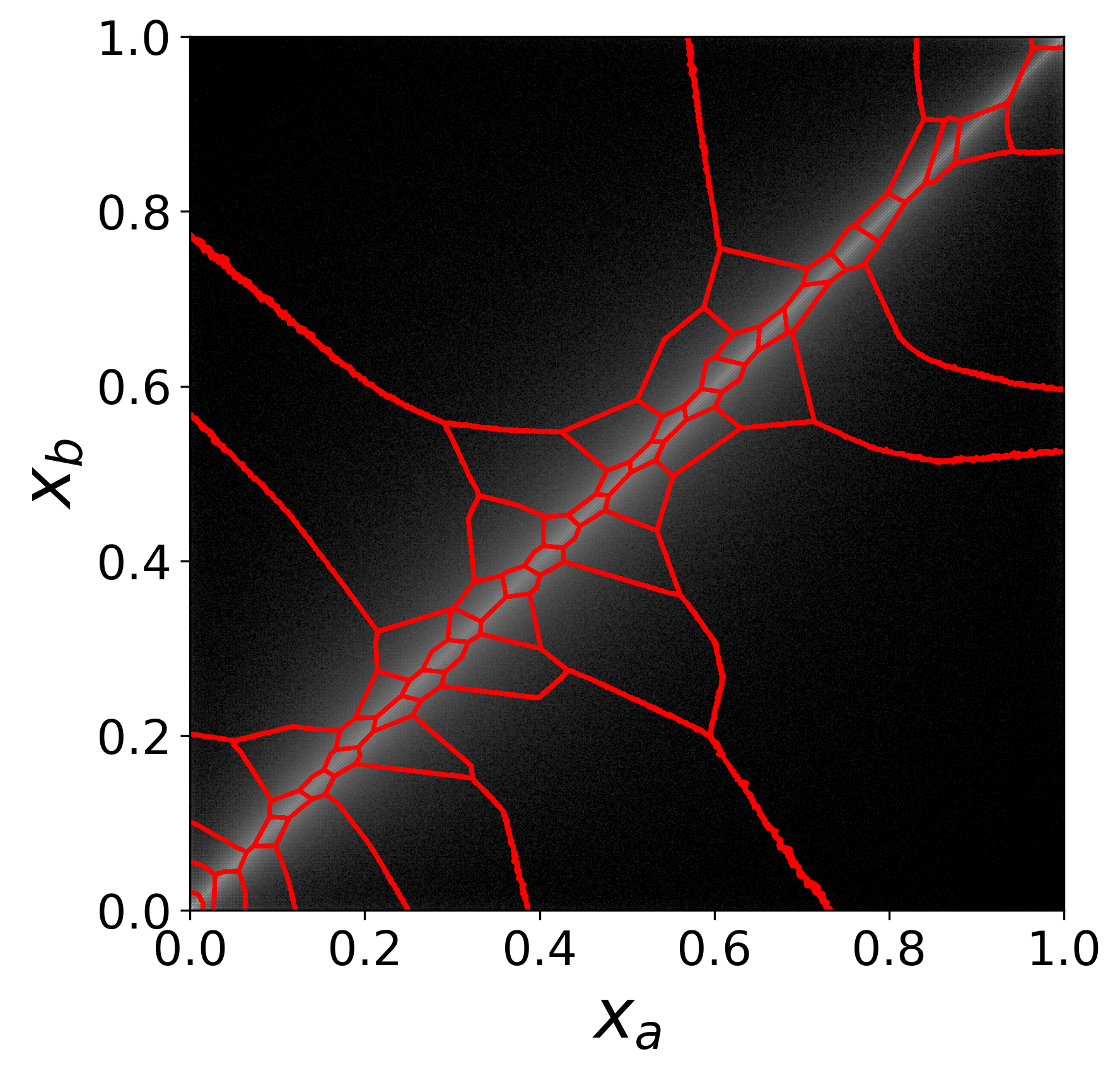} &  \includegraphics[width=0.45\textwidth]{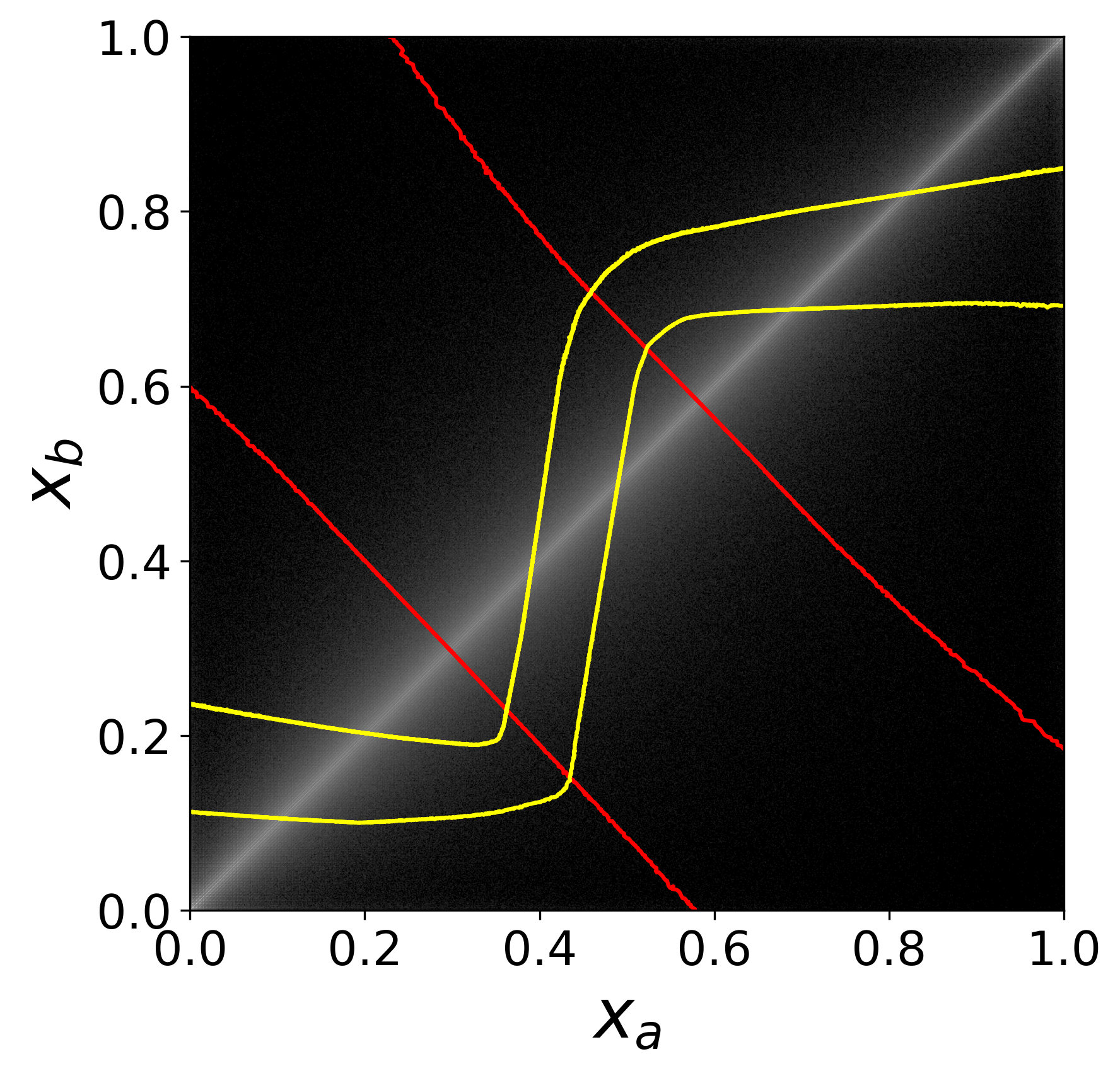}  \\
    (a)  & (b) \\
    \end{tabular}
    \caption{
        Additional results complementing Fig.~\ref{fig:2px_partitions} in the main text.
        (a) One output dimension with 64 states. The structure of the even partition is complex, 
            reflecting the high expressiveness of the MLP in modeling the probability space.
        (b) Two independent output dimensions, each with 3 states. 
            The second MLP has learned to employ a zigzag partitioning structure, effectively dividing the total intensity into more sections. 
            This suggests that the model may be compensating for the perceived insufficiency of the first MLP in partitioning the total intensity $x_a + x_b$
            with only 3 states.    
            }
\label{fig:2px_partitions2}
\end{figure}

In addition to Fig.~\ref{fig:2px_partitions} in the main text, some interesting results are shown in Fig.~\ref{fig:2px_partitions2}.
The partitions were visualized using the tricontour function from matplotlib.

\section{Experimental Details for the Image Patches Case}
\label{appendix:imagepatch_exp}

Two models were discussed in the main text of our study: a model trained on $5 \times 5$ color image patches (the color model), 
and another trained on $4 \times 4$ grayscale image patches (the grayscale model).

\textbf{Grayscale Model:} The training data consisted of roughly 100,000 images from the unlabeled section of the COCO 2017 image dataset, converted to grayscale. 
In each training batch, we randomly selected 1,000 images and extracted 1,000 random image patches from each image, leading to a total of 1 million image patches per batch. 
From this batch, a mini-batch of 500 patches was randomly chosen to calculate the loss function. 
To ensure numerical stability, a small value $\epsilon = 10^{-38}$ was added when computing the distances between samples. 
We used the Adam optimizer with a learning rate of 1e-3, and the model was trained for a single epoch. 
The sparsity regularization parameter was set to $\alpha=0.05$. 

\textbf{Color Model:} Training data for the color model were image patches extracted from 1.2 million images in the training portion of ImageNet. 
Each training batch comprised 500 randomly chosen images, with 100 random image patches extracted from each image. 
From each batch, a mini-batch of 500 patches was sequentially selected (i.e., patches within each batch were not shuffled) to compute the loss function. 
To augment the dataset, we randomly flipped images horizontally with a probability of 0.5. 
We adopted the node-wise loss function to ensure numerical stability (comparable results were achieved with the sample-wise loss function with $\epsilon = 10^{-38}$ added to the distance). 
The model was trained for 10 epochs using the AdamW optimizer, with a learning rate of 2e-4 for the first 5 epochs, and 1e-4 for the remaining epochs. 
The sparsity regularization parameter was set to $\alpha=\frac{6}{96}=0.0625$.

\section{Statistical Analysis of Even Code Representation}
\label{appendix:output_stat}
In this section, we conduct further statistical analyses on the even code representation using the grayscale model described in Appendix~\ref{appendix:imagepatch_exp}.
Fig.~\ref{fig:outputs2}~(a) illustrates the proportion of samples activating each unique binary representation.
At the most granular level, the distribution is highly uneven, reflecting the image similarity statistics learned by the model, as anticipated. 
Fig.~\ref{fig:outputs2}~(b) shows the proportion of samples activating a certain number of output nodes, 
with the majority activating around 14 out of the 64 total output nodes.

To confirm that samples also have a relatively even distribution at middle scales, 
We select 30 random occupied positions in the binary representation space and count the total number of samples within varying distances to each.
If the distribution is relatively even, the 30 curves should be close to each other (within the same order of magnitude) and exhibit similar shapes.
The result is shown in Fig.~\ref{fig:outputs2}~(c).
Most curves are close even at the smallest scale, and they become closer as the scale increases
In Fig.~\ref{fig:outputs2}~(d) we plot the occupancy rate of positions in the binary representation space at 
different distances to 30 random occupied positions. The occupancy rate is calculated as the total number of
samples at a distance $d$ to a random occupied position, divided by the number of all possible sites
in the binary space with a distance $d$ to the same position, which is the combination number ${64 \choose d}$.
A similar decaying trend is observed for all 30 random sites, except at a small scale or at a large $d$, where the boundary of the subspace is reached.

\begin{figure}
    \centering
    \begin{tabular}{cc}
       \includegraphics[width=0.45\textwidth]{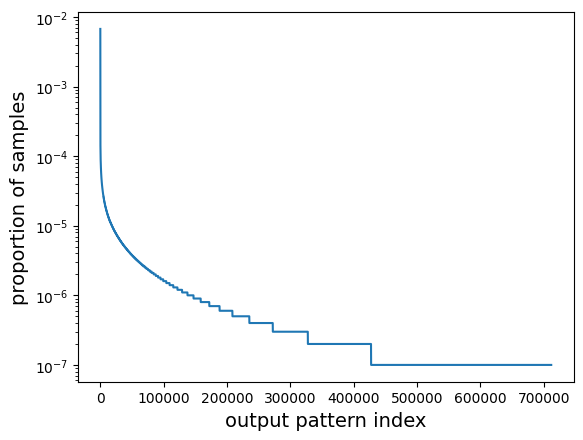} &   \includegraphics[width=0.45\textwidth]{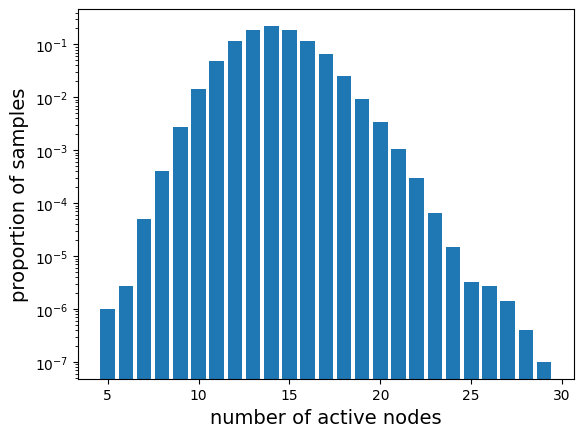} \\
      (a) & (b)  \\
      \includegraphics[width=0.45\textwidth]{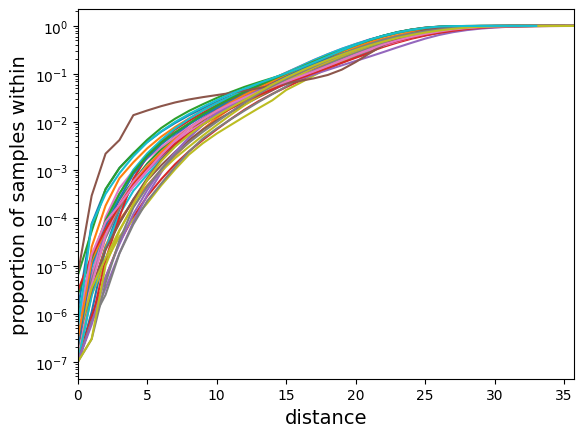} &   \includegraphics[width=0.45\textwidth]{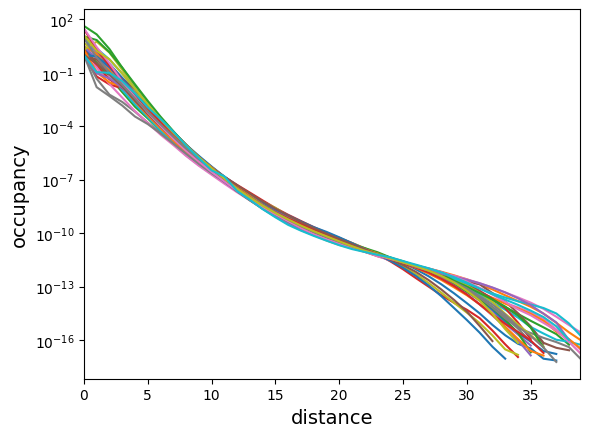} \\
      (c) & (d)  \\
      \end{tabular}
      \caption{Statistical analysis of the learned representation using the loss function
      Eq.~(\ref{eq:microscopic_loss}) defined in the main text. 
      (a) Proportion of image patch samples activating each unique representation pattern.
      (b) Proportion of image patch samples with representation having different numbers of active nodes.
      (c) Proportion of image patch samples within varying distances to 30 random positions in the binary representation space.
      (d) Occupancy rate at different distances to 30 random positions in the binary representation space.
      }
\label{fig:outputs2}
\end{figure}

\section{Decoding Even Code}
To intuitively understand how well the binary representation preserves information, 
we can decode the binary representations and compare them to the original image patches. 
The encoder model and its training is the same with the grayscale model described in Appendix~\ref{appendix:imagepatch_exp}
except here we use color images and $\alpha=0.03$.

To train the decoder, 10 million random image patches are fed into the encoder 
to generate corresponding representations. Since the mapping is many-to-one, 
we average all image patches with the same binary representation as the training target 
and use the binary vector as the feature. 
This results in approximately half a million feature-target pairs for training the decoder. 
We use an MLP with the same size as the encoder to construct the decoder, 
which is trained with a batch size of 128 for 100 epochs, using the Adam optimizer and a learning rate of 0.001.

\begin{figure}
    \centering
    \begin{tabular}{cc}
        \includegraphics[width=0.45\textwidth]{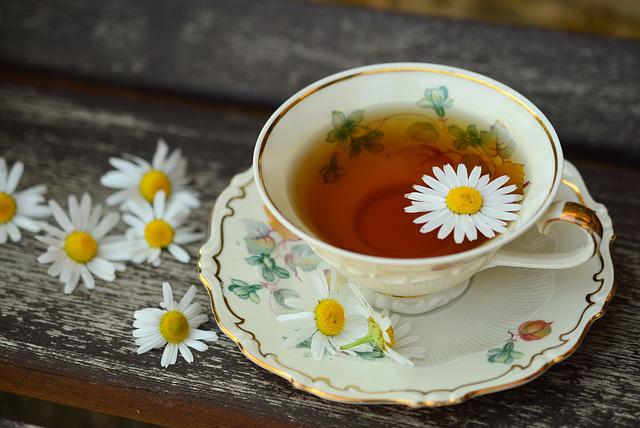} &   \includegraphics[width=0.45\textwidth]{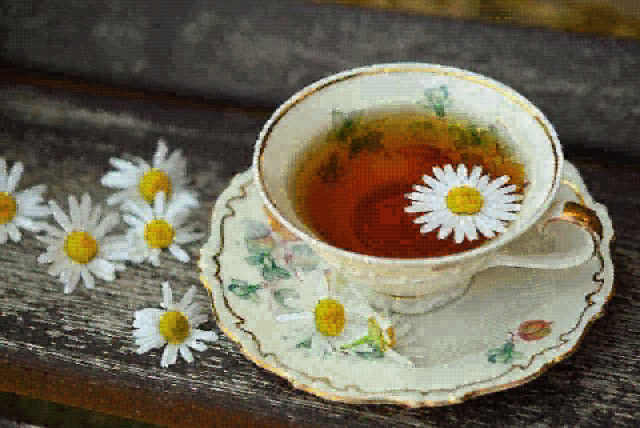} \\
      (a) & (b)
      \end{tabular}
      \caption{
        Demonstration of decoding the binary representation.  
        (a) The original image. 
        (b) The image reconstructed from decoded image patches, which are titled together. 
        Despite the lack of explicit optimization for image reconstruction, key details such as luminance, color, and boundaries are preserved effectively.
      }
\label{fig:decode}
\end{figure}

Once the decoder is trained, we first use the encoder to encode every non-overlapping image patch of an image. 
Then, the decoder is used to convert the binary representations back into image patches, 
which are then tilted together to get the decoded image.
Fig.~\ref{fig:decode} shows an example of original and decoded images. 
While the even code method does not explicitly optimize for image reconstruction performance in the loss function, 
unlike many previous image patch modeling methods \citep{Lee1999, ranzato2006efficient, NIPS2006_2d71b2ae, Osindero2008, Ranzato2010, Olshausen1996}, 
it is noteworthy that the binary representation preserves critical information such as luminance, color, and boundaries effectively. 
This retention of key details ensures the accurate comprehension of the image in subsequent processing stages.

\section{Complete output response to different gray levels and colors}
\label{appendix:color_luminance_all}
\begin{figure}
    \centering
    \includegraphics[width=0.9\textwidth]{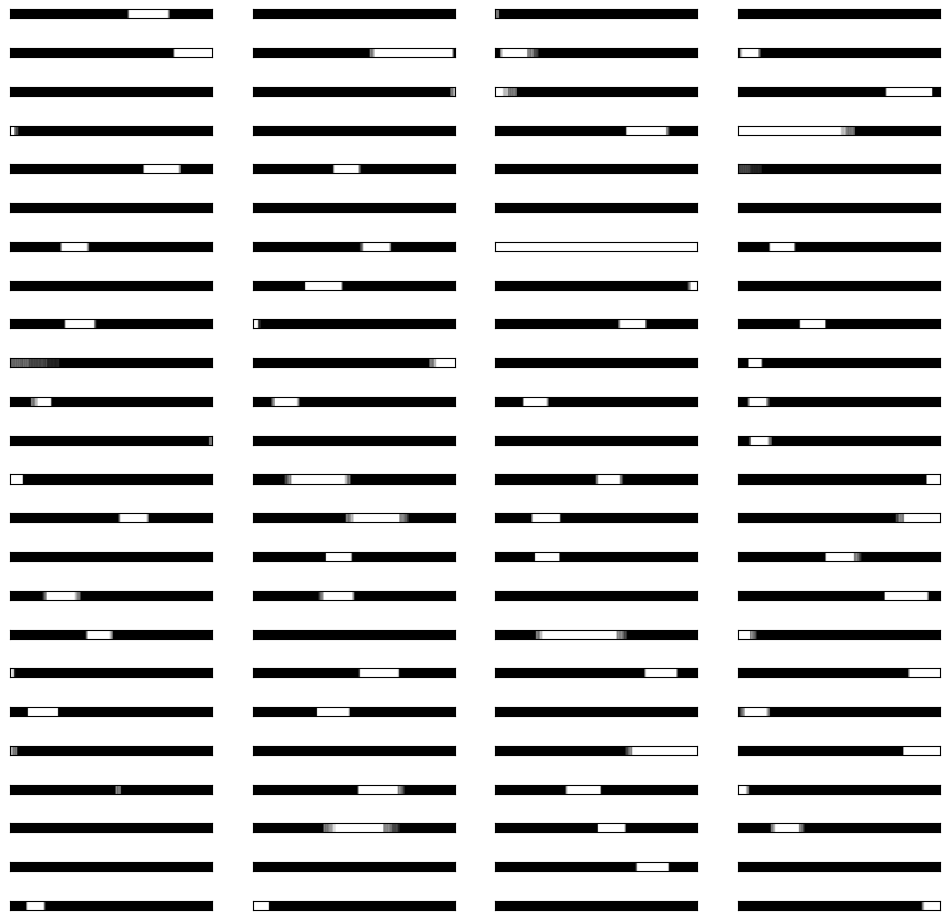} 
    \caption{Response of all 96 outputs from the $5 \times 5$ color IPU model to the linear grayscale input.}
\label{fig:gray_all}
\end{figure}

\begin{figure}
    \centering
    \includegraphics[width=0.9\textwidth]{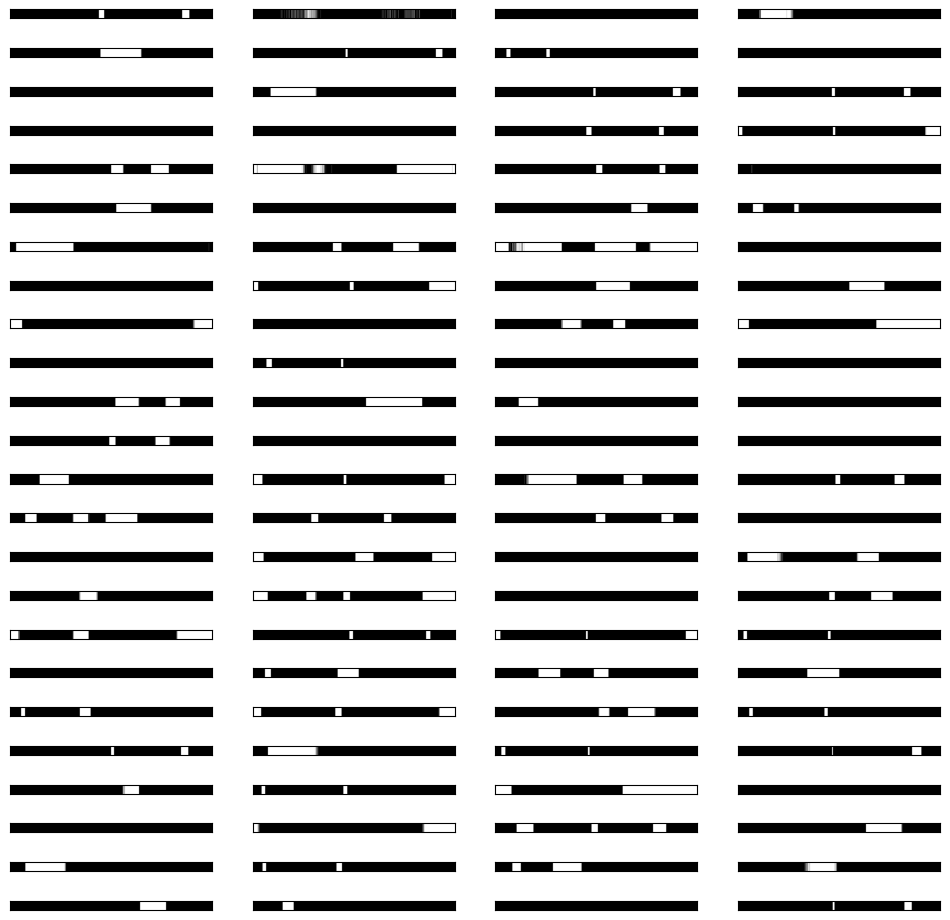} 
    \caption{Response of all 96 outputs from the $5 \times 5$ color IPU model to the spectral color input.}
\label{fig:color_all}
\end{figure}

Fig.~\ref{fig:gray_all} and Fig.~\ref{fig:color_all} display the responses of all 96 outputs from the $5 \times 5$ color IPU model 
to linear grayscale and spectral color inputs, as shown in Fig.~\ref{fig:color_luminance} in the main text.
Both input images have dimensions of 1920 by 90 pixels. 
The IPU processes the input images with a stride of 1 pixel, producing feature maps for each output node.

\end{document}